\documentclass[10pt,twocolumn,letterpaper]{article}

\usepackage[pagenumbers]{cvpr} 

\definecolor{cvprblue}{rgb}{0.21,0.49,0.74}
\usepackage[pagebackref,breaklinks,colorlinks,allcolors=cvprblue]{hyperref}

\usepackage{booktabs}
\usepackage{multirow}
\usepackage{graphicx}
\usepackage{amsthm}
\newtheorem{definition}{Definition}
\usepackage{pifont}

\usepackage{utfsym}
\usepackage[table]{xcolor}
\usepackage{makecell}
\usepackage{algorithmic}
\usepackage[ruled,vlined,linesnumbered]{algorithm2e}
\definecolor{stagecolor}{RGB}{0,150,160}
\definecolor{lightpurple}{RGB}{245,240,255}

\usepackage{capt-of}
\usepackage{subcaption}
\usepackage{bm}
\usepackage{tabularx}

\def\confName{CVPR}
\def\confYear{2027}

\title{Seeing a Watermark through Insects: Cavity Splatting in 3D Gaussian Scenes}

\author{Yang Chen$^{1,2}$\thanks{Work done while interning at Jilin University.~$^\dagger$Corresponding author.} ~ Qinrui Zhang$^1$ ~ Jiamin Yao$^1$ ~Yi Yu$^{1\dagger}$ ~ Chenqi Kong$^3$ ~ Yueqi Duan$^4$ ~ Yap-Peng Tan$^1$
\\
$^1$Jilin University ~\quad
$^2$Nanyang Technological University ~\quad
$^3$National University of Singapore \\
$^4$Tsinghua University ~\quad
$^5$VinUniversity
\\
{\tt\small chen1605@e.ntu.edu.sg ~\quad yuyi26@jlu.edu.cn}
}

\begin{document}
\maketitle
\begin{abstract}
What if seeing a watermark is a matter of where you stand? In nature, pigment cells in insect compound eyes optically isolate neighboring visual units, limiting light leakage. Inspired by the pigment sheath surrounding each visual unit, we introduce \textbf{Cavity Splatting}, a post-watermarking method constructing a tapered cavity aligned with target viewing rays. The cavity is a set of arranged Gaussians comprising an internal watermark and external sidewalls, from which we first develop boundary-aware placement of the cavity and design Perceptual Contrast Learning (PCL) to leave the watermark significantly visible through its aperture. Yet, applying this principle to a Gaussian scene introduces a challenge: the sidewalls that conceal the watermark can themselves become visible. To address this, we propose blending optimization (BO) to adaptively adjust the appearance of the sidewalls to blend with the surrounding backgrounds. Our method operates directly on scenes produced by feed-forward and optimization-based 3D Gaussian Splatting. Experiments on NeRF Synthetic, Mip-NeRF 360, RealEstate10K, and ACID show that our method controls view-selective visibility, preserving legibility in the target view and suppressing exposed artifacts in other views.

\end{abstract}    
\section{Introduction}
\label{sec:intro}

Advances in 3D Gaussian Splatting (3DGS)~\cite{kerbl20233d} have made 3D scenes increasingly accessible, creating a practical setting for sharing and marking reconstructed assets.
Conventional methods reconstruct scenes through scene-specific optimization~\cite{kerbl20233d, lu2024scaffold, chen2025dashgaussian, ren2026fastgs, yu2024mip}, while recent feed-forward methods directly predict Gaussian representations from sparse observations~\cite{zhang2025flare, zhang2026sparsesplat, zhang2026anchorsplat, jeong2026viewsplat, chen2026gifsplat, li2026tokensplat, hong2024pf3plat, liu2025monosplat}.
Representative methods~\cite{charatan2024pixelsplat, chen2024mvsplat} reduce reconstruction time while maintaining competitive rendering quality.
Building on this progress, geometric priors and relaxed camera-pose requirements further broaden the applicability of feed-forward reconstruction~\cite{xu2025depthsplat,kang2025selfsplat,long2026idesplat}.
These approaches produce 3D Gaussian assets that owners can share and viewers can explore from freely chosen viewpoints.


\begin{figure}[t]
  \centering
   \includegraphics[width=1.0\linewidth]{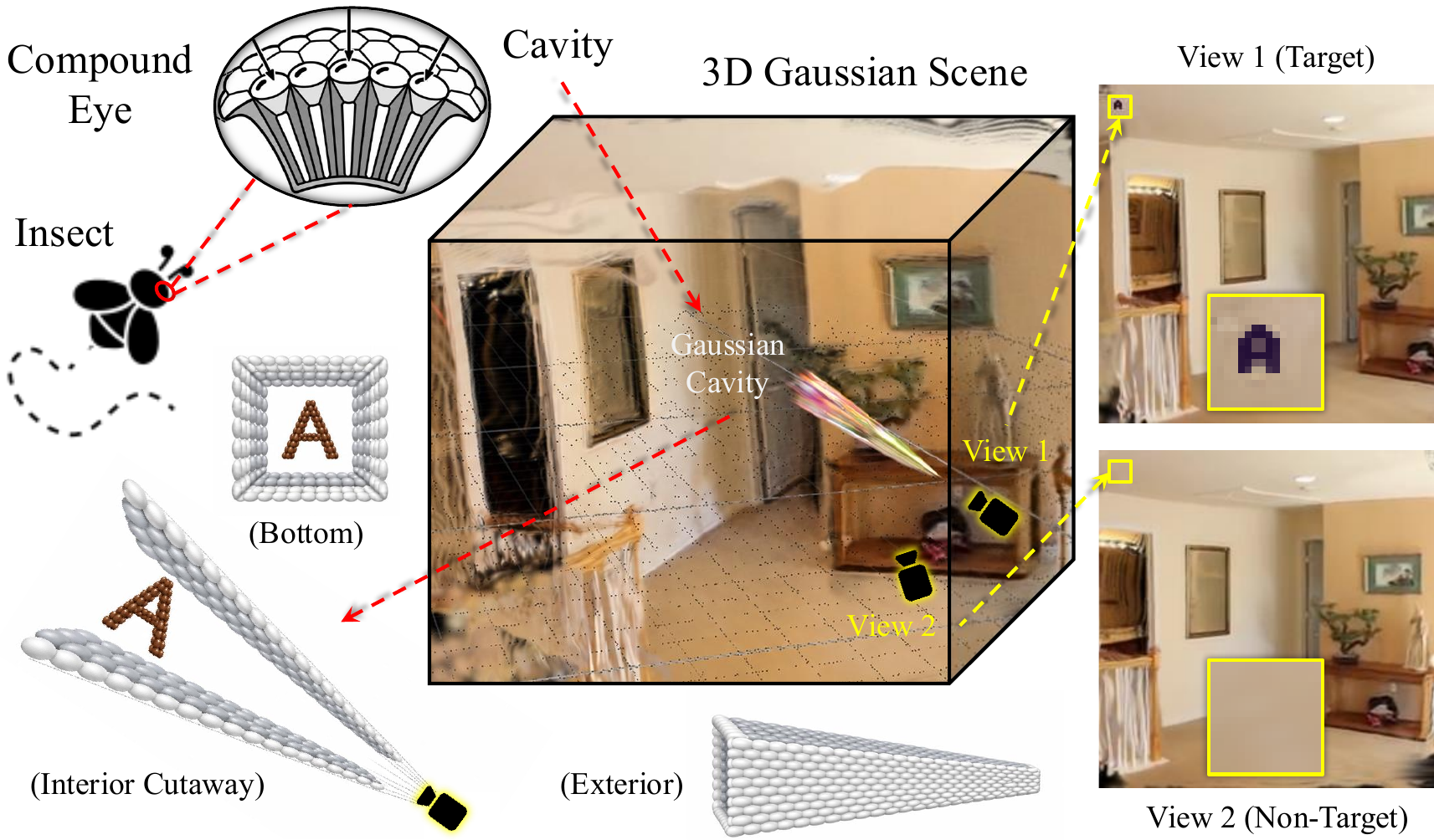}
   \caption{Cavity Splatting. Inspired by the compound eyes from insects (left), we place a watermark (glyph \texttt{A}) inside a tapered cavity with optimizable sidewalls comprised of Gaussians. Structural views show the cavity’s opening bottom, exterior, and interior with sidewalls omitted in the cutaway for clarity. Two cameras observe the same reconstructed room (center). The watermark is visible in the target view, while the neighboring view shows the room with the watermark concealed (right).}
   \label{fig:teaser}
\end{figure}

Sharing Gaussian scenes raises a practical question of attribution: \textit{a data owner may want a directly readable watermark identifying the source of a shared 3D asset}.
Watermarking has been widely studied for content authentication across images~\cite{fang2025syntag, qiu2025lightweight, qiu2026casial}, generative models~\cite{yang2024gaussianshading, yang2026fari}, 3D meshes~\cite{wang2022deep3d}, NeRF~\cite{luo2025nerfsignature, luo2026mantlemark}, and 3DGS~\cite{chen2025guardsplat, jang20253d, tan2024watergs, huang2024gaussianmarker, li2025gsmarker, huang2025marksplatter, li2026write}.
These methods mainly embed imperceptible messages for later decoding, leaving directly readable watermarking with controllable visibility across views less explored.

Consider a practical scenario in which an architectural studio shares a reconstructed apartment with prospective clients.
A small signature identifies the studio in a carefully composed overview, but could distract from the furnishings, layout, and materials as clients continue the walkthrough.
This tension between attribution and unobstructed exploration motivates view-selective visible watermarking: \textit{the owner may want a watermark to remain legible from a designated view while preserving scene appearance during ordinary exploration}.
Existing work has explored view-dependent visual illusions through NeRF and optimization-based 3DGS training~\cite{jiang2024ipa,ke2025stealthattack}, whereas the owner in this scenario already possesses a reconstructed asset.
We therefore study post-reconstruction visible watermarking, where embedding optimizes only added Gaussian primitives while keeping the original scene fixed.
Using the available scene and camera metadata, the owner selects target and non-target supervision views without requiring the original reconstruction images or the reconstruction model's parameters.
Operating directly on this Gaussian representation makes the embedding procedure generalizable to any 3DGS scenes, \textit{e.g.,} feed-forward and optimization-based scenes.

Inspired by a geometric intuition from insect compound eyes, where pigment surrounding each visual unit screens light entering through neighboring facets and helps keep their optical paths separate, we introduce \textbf{Cavity Splatting}. As illustrated in Fig.~\ref{fig:teaser}, the cavity forms an open passage via which the designated camera sees the watermark, while the neighboring camera views the scene with the watermark concealed. The structural views reveal the watermark inside the tapered cavity and the sidewalls along the viewing rays. To control watermark visibility, we design boundary-aware placement of the cavity and design Perceptual Contrast Learning (PCL) to leave the watermark significantly visible through its aperture, as well as designing blending optimization (BO) to adaptively adjust the appearance of sidewalls into the surrounding backgrounds. The method operates directly on an existing Gaussian representation and accommodates scenes produced by either feed-forward reconstruction or per-scene optimization. 
Experiments across four datasets and three reconstruction backbones achieve 91--100\% target-view recognition and 0--3.75\% non-target recognition for the glyph \texttt{A}, while maintaining good rendering fidelity.
A separate 26-letter evaluation demonstrates flexibility in watermark content, while evaluations against watermark removals show the robustness.

Our contributions are summarized as follows:



    

    
\begin{itemize}
    \item We formulate view-selective visible watermarking for already reconstructed Gaussian scenes, targeting a legible mark at a designated view while suppressing non-target exposure and preserving the original scene.

    \item Inspired by optical screening in insect compound eyes, Cavity Splatting integrates a ray-aligned Gaussian cavity with boundary-aware placement, PCL, and BO to coordinate geometric occlusion and appearance optimization, addressing watermark readability and sidewall artifacts.

    \item We evaluate the method on scenes from feed-forward and optimization-based 3DGS across four datasets, assessing target-view readability and non-target-view appearance.
\end{itemize}

\section{Related Work}
\label{sec:related_work}

\paragraph{3D Gaussian Splatting Watermarking.} 3D Watermarking representations have largely focused on imperceptible messages recovered from scene parameters or rendered images~\cite{huang2024gaussianmarker,jang20253d,wu2026fakemark,chen2025guardsplat,jang20253d,song2024geometrycloak}. Earlier NeRF methods~\cite{luo2025nerfsignature, song2024nerfprotector, luo2026mantlemark} encode binary messages in color representations~\cite{luo2023copyrnerf} or geometric components~\cite{huang2024geometrysticker}. For existing Gaussian scenes, GaussianMarker~\cite{huang2024gaussianmarker} freezes the original primitives and uses uncertainty-guided densification to introduce additional Gaussians for watermark optimization.
GuardSplat~\cite{chen2025guardsplat} optimizes
spherical-harmonic offsets, while 3D-GSW~\cite{jang20253d} combines scene fine-tuning with frequency-guided densification to improve invisible watermark robustness and rendering quality. Moving beyond per-scene watermark optimization, MarkSplatter~\cite{huang2025marksplatter} learns a generalizable embedding for single-pass message insertion.

\vspace{1mm}\noindent\textbf{View-Dependent 3D Rendering.}
In 3D representation learning~\citep{chen2024mvsplat, long2026idesplat, wang2024unlearnable, yan2026vfacamou, song2024geometrycloak}, view-dependent rendering aims to control what is observed from different views within a shared 3D representation. Illusion3D~\cite{feng2024illusion3d} uses pretrained 2D diffusion priors and differentiable rendering to optimize 3D textures and geometry. IPA-NeRF~\cite{jiang2024ipa} perturbs training images to induce a prescribed illusion at a backdoor viewpoint while preserving other views. StealthAttack~\cite{ke2025stealthattack} introduces
density-guided points into the initial Gaussian cloud and perturbs training views to induce viewpoint-dependent illusions. GaussTrap~\cite{hong2026gausstrap} exploits adaptive Gaussian density control and staged optimization to induce localized visual corruption while retaining fidelity in benign views. For copyright protection, Geometry Cloak~\cite{song2024geometrycloak} perturbs input
images so that unauthorized Gaussian reconstruction produces recognizable geometric patterns. Multi-view camouflage~\cite{owens2014camouflaging} provides a perspective on making added objects blend into their surroundings across cameras. 

\vspace{1mm}\noindent\textbf{Motivation and Novelty.}
Prior work mainly targets hidden messages or training-time illusions.
We enable view-selective visible watermarking in frozen Gaussian scenes through added occlusion geometry and appearance optimization, preserving readability and scene fidelity.

\section{Preliminaries and Problem Formulation}

\subsection{Gaussian Scenes and Rendering}
\label{sec:ew-main-preliminaries}

A reconstructed scene is represented by Gaussian primitives $\mathcal{G}
    \!=\! \{(\mu_j,\alpha_j,\Sigma_j,c_j)\}_{j=1}^{N}$,
where $\mu_j$, $\alpha_j$, $\Sigma_j$, and $c_j$ are the center, opacity, spatial covariance, and spherical-harmonic (SH) color coefficients, respectively.
The covariance is parameterized as
$\Sigma_j
    \!=\! Q_j \operatorname{diag}(s_{j1}^{2},s_{j2}^{2},s_{j3}^{2})Q_j^\top$,
where $Q_j$ is a rotation matrix and $s_j$ contains the principal-axis scales.

For a camera set $\mathcal{P}=\{P_k\}_{k=1}^{K}$, each camera is represented as $P_k=(K_k,R_k,t_k)$, with pixel-space intrinsics $K_k$ and a camera-to-world pose $(R_k,t_k)$ under the OpenCV camera-axis convention.
The clean rendering at $P_k$ is
\begin{equation}\small
    Y_k = \mathcal{R}(\mathcal{G},P_k),
    \quad k=1,\ldots,K,
    \label{eq:3dgs_encoder_decoder}
\end{equation}
where $\mathcal{R}$ projects the Gaussians and alpha-composites their contributions in depth order.

\subsection{Problem Formulation}
\label{sec:problem_formulation}

Our construction augments $\mathcal{G}$ with watermark Gaussians $\mathcal{W}_{\zeta}$ and sidewall Gaussians $\mathcal{C}_{\phi}$.
The parameters $\zeta$ control the watermark colors, while $\phi$ controls the sidewall geometry, opacity, and appearance.
The augmented rendering is
\begin{equation}\small
    \widehat{Y}_k(\zeta,\phi)
    = \mathcal{R}
    (\mathcal{G}\cup\mathcal{W}_{\zeta}\cup\mathcal{C}_{\phi},P_k).
    \label{eq:watermarked_rendering}
\end{equation}

\begin{definition}[View-Selective Visible Watermarking]
\label{def:view_selective_watermarking}
Given a Gaussian scene $\mathcal{G}$, a selected target camera $P_\star$, and a prescribed watermark glyph whose strokes occupy the pixel set $M$ in the target image, let $\mathcal{S}$ index the non-target cameras.
Ideally, non-target cameras cover all viewpoints other than $P_\star$.
Let $\mathcal{V}(Y,M)\in\{0,1\}$ indicate whether the watermark glyph specified by $M$ is visually recognizable in image $Y$.
The task seeks parameters $(\zeta^\star,\phi^\star)$ whose renderings ideally satisfy
\begin{equation}\small
    \mathcal{V}
    \bigl(\widehat{Y}_k(\zeta^\star,\phi^\star),M\bigr)
    =
    \begin{cases}
        1, & k=\star,\\
        0, & k\in\mathcal{S}.
    \end{cases}
    \label{eq:watermark_formulation}
\end{equation}
Meanwhile, $\widehat{Y}_\star$ should preserve $Y_\star$ outside $M$, and $\widehat{Y}_k$ should remain close to $Y_k$ for $k\in\mathcal{S}$ to suppress residual watermark and sidewall artifacts.
The resulting scene is
\begin{equation}\small
    \mathcal{G}_{\mathrm{wm}}
    = \mathcal{G}
    \cup \mathcal{W}_{\zeta^\star}
    \cup \mathcal{C}_{\phi^\star}.
    \label{eq:watermarked_scene}
\end{equation}
Only $\zeta$ and $\phi$ are optimized.
The original scene $\mathcal{G}$ and renderer $\mathcal{R}$ remain fixed throughout embedding.
\end{definition}

\subsection{Threat Model}
\label{sec:threat_model}

\textbf{Owner capabilities.}
A legitimate owner has access to the reconstructed scenes and cameras, and embeds the watermark before distribution.
Neither the original images nor the reconstruction model's parameters are required.

\vspace{1mm}\noindent\textbf{Recipient capabilities and protection scope.}
A recipient can render $\mathcal{G}_{\mathrm{wm}}$ from arbitrary cameras.
The owner privately retains $P_\star$ for watermark inspection without disclosing its exact parameters.
Furthermore, an adversarial recipient may also seek to remove or obscure the watermark while preserving scene quality, and 
we also examine such robustness against watermark removal methods, such as parameter perturbations, Gaussian merging, and pruning.

\vspace{1mm}\noindent\textbf{Practical scenario.}
An architectural visualization studio embeds its initial as a small watermark in a reconstructed apartment scene $\mathcal{G}$ before sharing it with prospective clients.
The studio privately retains $P_\star$, from which the mark is directly readable in an image corner with little disruption to the surrounding scene.
View-selective concealment reduces its exposure during walkthroughs, allowing clients to focus on the furnishings and layout.
If the watermarked asset $\mathcal{G}_{\mathrm{wm}}$ is later redistributed or reused, the studio can return to the retained view to reveal the embedded mark and support its copyright claim.

\section{Methodology}
\label{sec:method}

\subsection{Cavity Construction}
\label{sec:ew-main-cavity-construction}

\begin{figure}[t]
  \centering
   \includegraphics[width=1.0\linewidth]{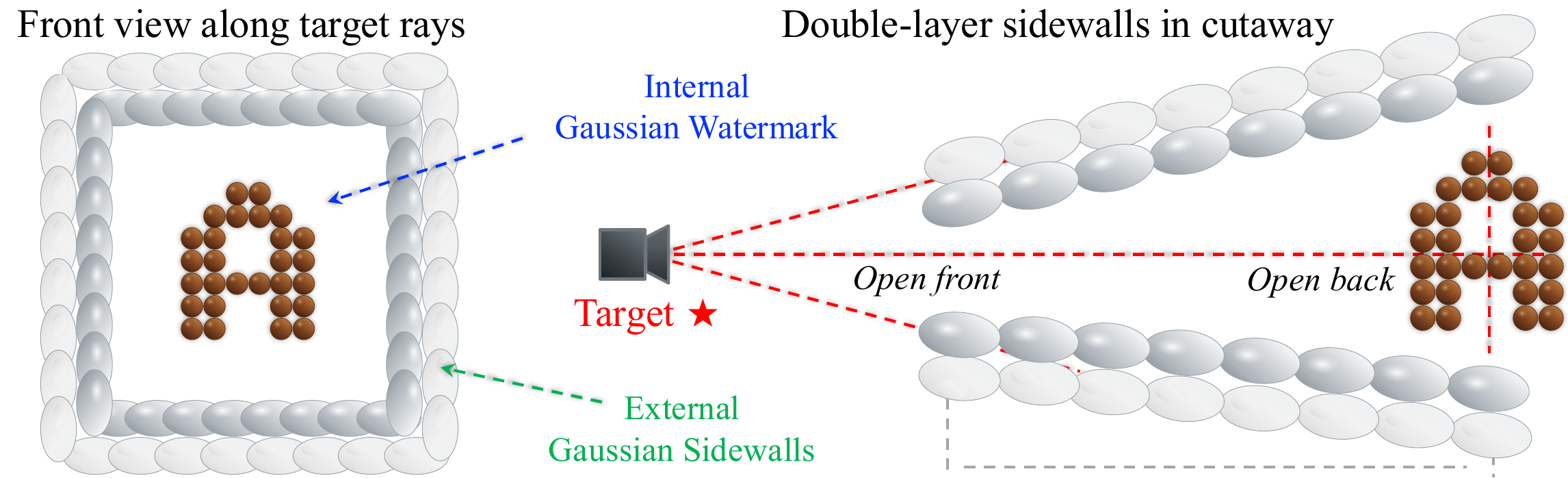}
   \vspace{-6mm}
   \caption{Cavity construction. The cavity consists of an internal Gaussian watermark and external Gaussian sidewalls.}
   \label{fig:cavity}
   \vspace{-4mm}
\end{figure}

As shown in Fig.~\ref{fig:cavity}, we construct a tapered Gaussian cavity to implement the view-selective visibility in Def.~\ref{def:view_selective_watermarking}.
We first select the target camera $P_\star$ and a compact watermark placement near the boundary of rendered scene coverage.
We then back-project stroke pixels $M$ onto a plane to instantiate the internal watermark $\mathcal{W}_\zeta$.
Finally, we surround the watermark with two layers of ray-aligned Gaussian sidewalls $\mathcal{C}_\phi$, forming an open passage along the target viewing rays while providing occlusion from off-axis directions.

\subsubsection{Boundary-Aware Placement}
\label{sec:ew-main-placement}

We seek a compact placement near the boundary of rendered scene coverage with limited exposure under reference cameras.
For each scene, let $\mathcal{P}_s^{\mathrm{data}}$ denote its available dataset cameras, with normalized focal lengths $\bar f_{a,k}^{\mathrm{crop}}$ computed under a common resizing and center-cropping convention.
To obtain a compact ray bundle for a fixed image-space glyph, we enlarge the maximum focal length independently along each axis:
\begin{equation}\small
    \bar f_a^\star
    = \kappa
    \max_{P_k\in\mathcal{P}_s^{\mathrm{data}}}
    \bar f_{a,k}^{\mathrm{crop}},
    \qquad a\in\{x,y\},
    \label{eq:ew-main-target-intrinsics}
\end{equation}
where $\kappa$ is the enlargement factor.
At target resolution $H\times W$, we set
$f_x^\star=W\bar f_x^\star$ and
$f_y^\star=H\bar f_y^\star$, center the principal point, and keep these intrinsics fixed during pose search.

Starting from an existing camera pose, we explore backward motion, corner-aligning rotations, and small lateral translations.
For each candidate $P$, we define
\begin{equation}\small
    \begin{aligned}
        U_P(p)
        &= \mathbf{1}\!\left[
            \alpha_{\mathcal{G}}(p,P)\leq\epsilon_\alpha
        \right],
        \quad N_P(D)
        = \sum_{p\in D}U_P(p),
    \end{aligned}
    \label{eq:ew-main-low-coverage}
\end{equation}
where $\alpha_{\mathcal{G}}(p,P)$ is the accumulated scene opacity, $N_P(D)$ counts low-coverage pixels in $D\!\subseteq\!\!\Omega$, and $\Omega$ is the image domain.
We select the corner window $Q$ maximizing $N_P(Q)$, then place the glyph within it to maximize $N_P(M_\xi)$, where $M_\xi$ is the stroke mask translated by offset $\xi$.

We further screen the proposed cavity and its viewing corridor using a heuristic bounding sphere $(b,r_C)$.
For each original Gaussian with $\alpha_j\geq\tau$, we bound its $\tau$-level ellipsoid by a sphere $(\mu_j,r_j)$ and compute
\begin{equation}\small
    d_{\mathrm{clear}}
    = \min_{j:\alpha_j\geq\tau}
    \left(
        \|b-\mu_j\|_2-r_j-r_C
    \right).
    \label{eq:ew-main-clearance}
\end{equation}
Let $q_{\mathrm{out}}$ be the fraction of reference cameras whose crop-adjusted frusta exclude the entire sphere, without the target focal enlargement.
A strict candidate satisfies
\begin{equation}\small
    \begin{gathered}
        N_P(Q)\geq n_{\min},\qquad
        N_P(\Omega\setminus Q)=0,\\
        N_P(M_\xi)\geq h_{\min},\qquad
        d_{\mathrm{clear}}>0,\qquad
        q_{\mathrm{out}}\geq q_{\min}.
    \end{gathered}
    \label{eq:ew-main-placement-criteria}
\end{equation}
The search stops at the first strict candidate.
If none is found within budget, it uses a ranked fallback prioritizing $N_P(\Omega\!\setminus\! Q)\!=\!0$.
This yields $P_\star$ and the placed stroke mask $M$.

\subsubsection{Internal Gaussian Watermark}
\label{sec:ew-main-internal-watermark}

Let $B$ be the selected $h\times w$ glyph box containing $M$.
We instantiate $\mathcal{W}_\zeta$ with one Gaussian per stroke pixel, giving $N_W=|M|$ primitives.
For a zero-based pixel index $p_j=(u_j,v_j)\in M$, we back-project its center onto a plane at target-camera depth $z_W$:
\begin{equation}\small
    \begin{aligned}
        q_j
        = K_\star^{-1}
        [u_j+\tfrac12,\ v_j+\tfrac12,\ 1]^\top,
        ~\mu_j^{\mathcal{W}}
        = t_\star+R_\star z_Wq_j.
    \end{aligned}
    \label{eq:ew-main-watermark-backprojection}
\end{equation}
Since $q_j$ has unit third component, $z_W$ gives the depth along the target camera's optical axis.
The Gaussians are aligned with the target-camera image axes.
For a given projected standard deviation $\sigma_p$, their lateral scales are initialized as
\begin{equation}\small
    s_x=\frac{z_W\sigma_p}{f_x^\star},
    \qquad
    s_y=\frac{z_W\sigma_p}{f_y^\star},
    \label{eq:ew-main-watermark-scales}
\end{equation}
with a smaller positive scale along the plane normal.
We verify the projected centers against their stroke pixels using the renderer.
The geometry and opacities remain fixed throughout embedding, while PCL learns the colors $\zeta$.

\subsubsection{External Gaussian Sidewalls}
\label{sec:ew-main-external-sidewalls}

We surround the watermark with two nested layers of Gaussian sidewalls.
Each layer has four walls extending from $z_{\mathrm{front}}$ to $z_{\mathrm{back}}=z_W$, with an open front and no rear cap.
The aperture extends beyond $B$ by a margin $m_{\mathrm{ap}}$, and the layers are separated by $\delta_{\mathrm{layer}}$.
Discretizing each wall into $n_{\mathrm{edge}}$ edge cells and $n_{\mathrm{depth}}$ depth cells gives
$N_C=8n_{\mathrm{edge}}n_{\mathrm{depth}}$ Gaussians.

\emph{1) Sidewall centers} are placed along rays framing the aperture at their assigned depths, following Eq.~\eqref{eq:ew-main-watermark-backprojection}.
At initialization, one principal axis of each Gaussian follows its target-camera ray to limit its transverse image footprint, and the wall-normal scale is multiplied by $\gamma_{\mathrm{normal}}$.
During optimization, the Gaussian count and local orientation frames remain fixed.
Centers stay within their initial cells, with the frontmost and rearmost layers retaining their target-camera depths.
Scales and opacities are bounded without truncating Gaussian tails.
\emph{2) Sidewall appearance} uses SH degree $L_{\mathrm{SH}}$ with $n_{\mathrm{SH}}=(L_{\mathrm{SH}}+1)^2$ coefficients per color channel.
The RGB color evaluated in direction $d$ is
\begin{equation}\small
    \rho_j(d)
    = \tfrac12
    + \sum_{q=0}^{n_{\mathrm{SH}}-1}
    c_{jq}\psi_q(d),
    \label{eq:ew-main-sh-color}
\end{equation}
where $\psi_q$ is an SH basis function.
Let $d_{j\star}$ be the renderer's viewing direction for Gaussian $j$ at $P_\star$.
We optimize a base RGB color $\bar c_j$ and the higher-order SH coefficients, deriving the DC coefficient via target-direction compensation:
\begin{equation}\small
    c_{j0}
    =
    \frac{
        \bar c_j-\tfrac12
        -\sum_{q=1}^{n_{\mathrm{SH}}-1}
        c_{jq}\psi_q(d_{j\star})
    }{\psi_0}.
    \label{eq:ew-main-sh-compensation}
\end{equation}
The direction and compensation are recomputed from the current center, with gradients propagated through both.
Before clipping, this gives $\rho_j(d_{j\star})=\bar c_j$.
Thus, higher-order coefficients adapt non-target appearance while $\bar c_j$ controls the target-direction color.
The optimized parameters are
\begin{equation}\small
    \phi
    =
    \left\{
        \mu_j^{\mathcal{C}},
        s_j^{\mathcal{C}},
        \alpha_j^{\mathcal{C}},
        \bar c_j,
        c_{j,1:n_{\mathrm{SH}}-1}
    \right\}_{j=1}^{N_C}.
    \label{eq:ew-main-cavity-parameters}
\end{equation}
BO in \Cref{sec:ew-main-blending} also supervises the composited target image to account for changes in geometry, opacity, and $\bar c_j$.

For later optimization, we use the augmented rendering defined above and the watermark-only target rendering:
\begin{align}
    \widehat{Y}_k(\zeta,\phi)
    &= \mathcal{R}
    (\mathcal{G}\cup\mathcal{W}_\zeta\cup\mathcal{C}_\phi,P_k),
    \label{eq:ew-main-augmented-rendering}\\
    \widetilde{Y}_\star(\zeta)
    &= \mathcal{R}
    (\mathcal{G}\cup\mathcal{W}_\zeta,P_\star).
    \label{eq:pco-rendering}
\end{align}
PCL in \Cref{sec:perceptual-contrast} learns $\zeta$ using $\widetilde{Y}_\star$, after which
$\mathcal{W}_{\zeta^\star}$ is frozen for sidewall optimization.

\begin{algorithm}[t]\small
\DontPrintSemicolon
\caption{Optimization of Cavity Splatting}
\label{alg:cavity-splatting}

\KwIn{Scene $\mathcal G$, cameras $\mathcal P_s^{\rm data}$,
glyph bitmap}
\KwOut{Watermarked scene $\mathcal G_{\rm wm}$,
target camera $P_\star$}

Select $P_\star$ and stroke mask $M$; Initialize $\mathcal W_\zeta$ with $\mathcal G$ fixed\;

\textcolor{stagecolor}{
\# Stage-1: PCL for internal Gaussian watermark}\;
\For{$t=1$ \KwTo $T_{\rm PCL}$}{
    Render $\widetilde Y_\star(\zeta)$
    with Eq.~\eqref{eq:pco-rendering}\;
    Update only $\zeta$ using
    $\mathcal L_{\rm PCL}$ in Eq.~\eqref{eq:pco-loss}\;
}
Retain the lowest-loss color state as $\zeta^\star$ and
\textbf{freeze} $\mathcal W_{\zeta^\star}$\;

Initialize $\mathcal C_\phi$ and select supervision views
$\mathcal S$ and clean renderings $Y_k$\;
Construct $J_\star$ with
Eq.~\eqref{eq:ew-main-target-image}\;

\textcolor{stagecolor}{
\# Stage-2: BO for external Gaussian sidewalls}\;
Calibrate sidewall colors against
$\widetilde Y_\star(\zeta^\star)$ \;
Record initial supports $\Omega_k^{(0)}$\;

\For{$t=1$ \KwTo $T_{\rm BO}$}{
    Render supervised views with
    Eq.~\eqref{eq:ew-main-augmented-rendering}\;
    Update $D_k^{(t)}$ with
    Eq.~\eqref{eq:ew-main-dynamic-region}\;
    Update only $\phi$ using
    $\mathcal L_{\rm BO}$ in Eq.~\eqref{eq:ew-main-objective}\;
}

Set $\phi^\star\leftarrow\phi$\;
$\mathcal G_{\rm wm}\leftarrow
\mathcal G\cup\mathcal W_{\zeta^\star}
\cup\mathcal C_{\phi^\star}$\;
\Return{$\mathcal G_{\rm wm}$, $P_\star$}\;

\end{algorithm}

\subsection{Cavity Optimization}
\label{sec:ew-main-cavity-optimization}

We pursue the readability and appearance preservation of Def.~\ref{def:view_selective_watermarking} in 2 stages, as in Alg.~\ref{alg:cavity-splatting}.
PCL first learns watermark colors that distinguish the glyph from its local background.
Fixing the learned watermark provides a stable target appearance for BO, which optimizes the sidewalls to suppress non-target exposure while preserving this appearance.

\subsubsection{Perceptual Contrast Learning (PCL)}
\label{sec:perceptual-contrast}

We initialize the watermark colors using the mean RGB of the clean target image $Y_\star$.
To make the compact glyph recognizable from its few stroke pixels, PCL encourages both luminance contrast and perceptual color separation from the local background.
Let $\Pi=M\times B$ contain all foreground--background comparison pairs.
For $(p,q)\in\Pi$, we compare the rendered stroke color
$f_p=\widetilde{Y}_\star(\zeta)(p)$ with the clean background color
$b_q=Y_\star(q)$, clipping both to $[0,1]$ before luminance and Lab conversion:
\begin{equation}\small
    \begin{aligned}
        CR_{pq}
        &=
        \frac{
            \max\{L_{\mathrm{rel}}(f_p),L_{\mathrm{rel}}(b_q)\}
            +\epsilon_{\mathrm{lum}}
        }{
            \min\{L_{\mathrm{rel}}(f_p),L_{\mathrm{rel}}(b_q)\}
            +\epsilon_{\mathrm{lum}}
        },\\
        \Delta E_{00}^{pq}
        &=
        \operatorname{CIEDE2000}
        \bigl(
            \operatorname{Lab}(f_p),
            \operatorname{Lab}(b_q)
        \bigr),
    \end{aligned}
    \label{eq:pco}
\end{equation}
where $L_{\mathrm{rel}}$ computes relative luminance from linearized sRGB, $\operatorname{Lab}$ converts RGB colors to CIELAB, and $\epsilon_{\mathrm{lum}}$ is the luminance offset.
We update only $\zeta$ using
\begin{equation}\small
    \begin{aligned}
        \mathcal{L}_{\mathrm{PCL}}
        \!=\!
        \frac{\lambda_{\mathrm{cr}}}{|\Pi|}
        \sum_{(p,q)\in\Pi}\!
        \left[
            \frac{\tau_{\mathrm{cr}}\!-\!CR_{pq}}
                 {\tau_{\mathrm{cr}}}
        \right]_{+}^{2}
        \!+\!
        \frac{\lambda_{\mathrm{de}}}{|\Pi|}
        \sum_{(p,q)\in\Pi}\!
        \frac{1}{1\!+\!\Delta E_{00}^{pq}},
    \end{aligned}
    \label{eq:pco-loss}
\end{equation}
where $[\cdot]_+=\max(0,\cdot)$ and $\tau_{\mathrm{cr}}$ is the desired contrast ratio.
The two terms penalize insufficient contrast and encourage perceptual color separation, weighted by $\lambda_{\mathrm{cr}}$ and $\lambda_{\mathrm{de}}$, respectively.
We retain the lowest-loss color state as $\zeta^\star$, freeze $\mathcal{W}_{\zeta^\star}$, and construct the target supervision image
\begin{equation}\small
    J_\star(p)
    =
    \begin{cases}
        \widetilde{Y}_\star(\zeta^\star)(p), & p\in M,\\
        Y_\star(p), & p\notin M.
    \end{cases}
    \label{eq:ew-main-target-image}
\end{equation}

\subsubsection{Blending Optimization (BO)}
\label{sec:ew-main-blending}

With $\mathcal{W}_{\zeta^\star}$ fixed, we initialize the sidewalls and write
$\widehat{Y}_k(\phi)\equiv\widehat{Y}_k(\zeta^\star,\phi)$.
BO selects non-target supervision views, calibrates the sidewall colors, and optimizes their geometry, opacity, and appearance.

\vspace{1mm}\noindent\textbf{Supervision views.}
We prioritize views with substantial appearance changes while maintaining diverse viewing conditions.
We generate non-target candidates by translating the target pose and orbiting around the scene center.
We retain candidates with nonempty added-structure projections and a maximum absolute RGB change exceeding $\epsilon_{\mathrm{RGB}}$.
Their indices form $\mathcal{I}_{\mathrm{cand}}$.
For each candidate $k$, the perturbation score $E_k$ measures the squared RGB difference between the augmented and clean renderings, summed over pixels and averaged over color channels.
These scores are computed with the learned watermark and initial sidewalls before sidewall color calibration. We define candidate diversity as
\begin{equation}\small
\begin{aligned}
D(k,\ell)
=w_t\widehat d_t(k,\ell)
 +w_R\widehat d_R(k,\ell)+w_P\left[1-\operatorname{IoU}(A_k,A_\ell)\right],
\end{aligned}
\label{eq:ew-view-diversity}
\end{equation}
\noindent
where $\widehat d_t$ and $\widehat d_R$ are the Euclidean
distance between camera centers and the relative rotation
angle on $\mathrm{SO}(3)$, respectively, each normalized
by its maximum over all retained candidate pairs.
The mask $A_k$ is the union of independently rasterized
watermark and sidewall supports at camera $k$.
For a nonempty candidate set, we initialize $\mathcal{S}$ with the camera having the largest $E_k$ and add
\begin{equation}\small
    k_{\mathrm{next}}
    =
    \arg\!\!\!\!\!\max_{k\in\mathcal{I}_{\mathrm{cand}}\setminus\mathcal{S}}
    [
        \lambda\sqrt{\tfrac{E_k}{E_{\max}}}
        +(1\!-\!\lambda)
        \min_{\ell\in\mathcal{S}}D(k,\ell)
    ],
    \label{eq:ew-view-selection}
\end{equation}
where $E_{\max}=\max_{k\in\mathcal{I}_{\mathrm{cand}}}E_k$ and $\lambda$ balances perturbation and diversity.
We select at most $K_{\mathrm{train}}$ cameras, using all retained candidates if fewer are available.
The set $\mathcal{S}$ remains fixed throughout BO.
We cache the clean renderings and use $P_\star$ and every selected non-target camera at each optimization step.

\vspace{1mm}\noindent\textbf{Sidewall color calibration.}
Before jointly optimizing the sidewall parameters, we calibrate the target-direction RGB colors of both layers while keeping their geometry and opacities fixed.
Calibration minimizes reconstruction error against the full watermark-only rendering $\widetilde{Y}_\star(\zeta^\star)$ over the sidewalls' target-view projection.

\vspace{1mm}\noindent\textbf{Appearance optimization.}
We concentrate appearance supervision on regions affected by the compact cavity.
Let $\Omega_k^{(0)}$ denote the standalone projection support of all added Gaussians after calibration, and let $\Omega_k^{(t)}$ denote their support at BO iteration $t$.
We define the optimization region as
\vspace{-1mm}
\begin{equation}\small
    \begin{aligned}
        D_k^{(t)}
        = \operatorname{stopgrad}\bigl(
            \Omega_k^{(0)}
            \cup \Omega_k^{(t)}
            \cup
            \{p:\widehat{Y}_k^{(t)}(p)\neq Y_k(p)\}
        \bigr),
    \end{aligned}
    \label{eq:ew-main-dynamic-region}
\end{equation}
where $\widehat{Y}_k^{(t)}=\widehat{Y}_k(\phi^{(t)})$.
We additionally include $M$ in the target-view region.
Projection supports are computed with the renderer, and the difference set includes any nonzero RGB change.
The region is recomputed as the geometry changes but excluded from backpropagation.
Keeping both initial and current supports accounts for areas affected by sidewall motion.
For a nonempty pixel region $D$, define
\vspace{-1mm}
\begin{equation}\small
    \ell_D(U,V)
    =
    \frac{1}{3|D|}
    \sum_{p\in D}\|U(p)-V(p)\|_2^2,\quad\text{with}~\ell_\varnothing=0.
    \label{eq:ew-main-masked-error}
\end{equation}
Suppressing iteration indices, the target and non-target losses are
\vspace{-1mm}
\begin{equation}\small
    \begin{aligned}
        \mathcal{L}_{\mathrm{tar}}
        &=
        \ell_M(\widehat{Y}_\star,J_\star)
        +\lambda_{\mathrm{out}}
        \ell_{D_\star\setminus M}(\widehat{Y}_\star,Y_\star),\\
        \mathcal{L}_{\mathrm{non}}
        &=
        \frac{1}{|\mathcal{S}|}
        \sum_{k\in\mathcal{S}}
        \ell_{D_k}(\widehat{Y}_k,Y_k),
    \end{aligned}
    \label{eq:ew-main-nontarget-loss}
\end{equation}
where $\mathcal{L}_{\mathrm{non}}=0$ when $\mathcal{S}=\varnothing$.
We update $\phi$ by minimizing
\vspace{-1mm}
\begin{equation}\small
    \mathcal{L}_{\mathrm{BO}}(\phi)
    =
    \lambda_{\mathrm{tar}}\mathcal{L}_{\mathrm{tar}}
    +\lambda_{\mathrm{non}}\mathcal{L}_{\mathrm{non}}.
    \label{eq:ew-main-objective}
\end{equation}
The stroke term preserves the target appearance learned by PCL, while the background and non-target terms suppress image changes introduced by the added structure.

\begin{table*}[t]
\centering
\caption{Recognition accuracy (\%) across models and datasets.
$\mathbf{A}$: watermark only;
$\mathbf{B}$: watermark with initialized, color-calibrated sidewalls;
$\mathbf{C}$: watermark with optimized sidewalls.
Seen and Unseen denote non-target views included in and excluded
from sidewall optimization, respectively.
Each group uses identical cameras across the three stages.
Higher is better for Target (\textcolor{purple}{$\uparrow$}),
while lower is better for Seen and Unseen
(\textcolor{stagecolor}{$\downarrow$}).
The best values within each $\mathbf{A/B/C}$ triplet are highlighted in the corresponding arrow color.}
\label{tab:exp-recognition}
\vspace{-3mm}

\footnotesize
\setlength{\tabcolsep}{2pt}
\renewcommand{\arraystretch}{1.05}

\resizebox{\textwidth}{!}{%
\begin{tabular}{@{}l|*{5}{ccc|}ccc@{}}
\toprule
\multirow{2}{*}{\shortstack{View\\group}}
& \multicolumn{3}{c|}{MVSplat / RE10K}
& \multicolumn{3}{c|}{MVSplat / ACID}
& \multicolumn{3}{c|}{IDESplat / RE10K}
& \multicolumn{3}{c|}{IDESplat / ACID}
& \multicolumn{3}{c|}{3DGS / MIP}
& \multicolumn{3}{c}{3DGS / NS} \\
\cmidrule(lr){2-4}
\cmidrule(lr){5-7}
\cmidrule(lr){8-10}
\cmidrule(lr){11-13}
\cmidrule(lr){14-16}
\cmidrule(lr){17-19}
& $\mathbf{A}$ & $\mathbf{B}$ & $\mathbf{C}$
& $\mathbf{A}$ & $\mathbf{B}$ & $\mathbf{C}$
& $\mathbf{A}$ & $\mathbf{B}$ & $\mathbf{C}$
& $\mathbf{A}$ & $\mathbf{B}$ & $\mathbf{C}$
& $\mathbf{A}$ & $\mathbf{B}$ & $\mathbf{C}$
& $\mathbf{A}$ & $\mathbf{B}$ & $\mathbf{C}$ \\
\midrule

Target (\textcolor{purple}{$\uparrow$})
& \textcolor{purple}{95.36} & 95.21 & \textcolor{purple}{95.36}
& 97.50 & 97.50 & \textcolor{purple}{98.75}
& 91.00 & \textcolor{purple}{92.00} & 91.00
& 96.25 & \textcolor{purple}{97.50} & 96.25
& \textcolor{purple}{100.00}
& \textcolor{purple}{100.00}
& \textcolor{purple}{100.00}
& \textcolor{purple}{100.00} & 87.50
& \textcolor{purple}{100.00} \\

Seen (\textcolor{stagecolor}{$\downarrow$})
& 74.65 & 2.78 & \textcolor{stagecolor}{1.24}
& 76.88 & 6.88 & \textcolor{stagecolor}{3.75}
& 76.00 & 5.00 & \textcolor{stagecolor}{1.00}
& 72.50 & 5.63 & \textcolor{stagecolor}{2.50}
& 73.61 & 6.94 & \textcolor{stagecolor}{2.78}
& 50.00 & 1.56 & \textcolor{stagecolor}{0.00} \\

Unseen (\textcolor{stagecolor}{$\downarrow$})
& 53.66 & 1.11 & \textcolor{stagecolor}{0.75}
& 30.10 & 0.42 & \textcolor{stagecolor}{0.31}
& 35.33
& \textcolor{stagecolor}{0.67}
& \textcolor{stagecolor}{0.67}
& 25.52 & \textcolor{stagecolor}{0.10} & 0.21
& 0.92
& \textcolor{stagecolor}{0.00}
& \textcolor{stagecolor}{0.00}
& 56.25 & \textcolor{stagecolor}{0.00} & 3.13 \\

\bottomrule
\end{tabular}%
}
\vspace{-5mm}
\end{table*}

\section{Experiments}
\label{sec:exp}

\subsection{Experimental Setup}

\textbf{Datasets \& Metrics.}
We evaluate feed-forward scenes from RealEstate10K
(RE10K)~\cite{zhou2018stereo} and ACID~\cite{liu2021infinite},
and optimization-based scenes from NeRF-Synthetic
(NS)~\cite{mildenhall2021nerf} and Mip-NeRF 360
(MIP)~\cite{barron2022mip}.
We measure distinct properties:
\textbf{recognition accuracy (acc.)} using an MLLM-as-a-judge (details in the appendix),
and full-image and local-region \textbf{fidelity} to the
original clean rendering using PSNR, SSIM, and LPIPS.
We aim for high target-view recognition, low recognition in nearby non-target views, minimal watermark and sidewall exposure in dataset views, and high local fidelity and robustness to watermark removals.


\vspace{1mm}\noindent\textbf{Reconstruction Backbones \& Evaluation States.}
The backbones are 3DGS~\cite{kerbl20233d}, MVSplat~\cite{chen2024mvsplat},
and IDESplat~\cite{long2026idesplat}. State $\mathbf{A}$ contains the learned watermark without sidewalls. State $\mathbf{B}$ adds initialized, color-calibrated sidewalls before BO, and state $\mathbf{C}$ uses the sidewalls after BO. The original scene and learned watermark remain identical across $\mathbf{A/B/C}$. \textbf{Target} denotes the camera selected by boundary-aware placement. \textbf{Seen} and \textbf{Unseen} denote local non-target cameras included in and excluded
from optimization. 

\vspace{1mm}\noindent\textbf{Implementations.}
Local rendering uses $256\!\times\!256$ images, an $8\!\times\!8$ glyph, and 512 sidewall Gaussians.
The main glyph \texttt{A} has 28 active pixels, yielding 28 watermark Gaussians and 540 added Gaussians.
Target focal lengths are $\kappa\!=\!2.5$ times the per-scene maxima of crop-adjusted normalized focal lengths along each axis, with normalized principal point $(0.5,0.5)$.
Placement uses a 120-second budget with $\epsilon_\alpha\!=\!\tau\!=\!0.01$, $n_{\min}\!=\!h_{\min}\!=\!4$, and $q_{\min}\!=\!0.95$.
We use absolute target-camera forward depths $(z_{\rm front},z_W,z_{\rm back})\!=\!(0.21,0.45,0.45)$, $\sigma_p=0.33$ pixels, and sidewall SH degree 3.
PCL optimizes watermark colors for $T_{\rm PCL}\!=\!50$ Adam steps at learning rate of 0.02, with $\lambda_{\rm cr}\!=\!\lambda_{\rm de}\!=\!1$ and $\tau_{\rm cr}\!=\!4.5$, the WCAG 2.0 minimum contrast ratio for normal text~\cite{caldwell2008web}.
We freeze the lowest-loss color state across initialization and all steps.
We select up to $K_{\rm train}\!=\!8$ fixed non-target cameras with diversity weights $(w_t,w_R,w_P)\!=\!(0.4,0.3,0.3)$ and perturbation--diversity weight $\lambda\!=\!0.5$.
After sidewall color calibration when needed, BO updates only the sidewalls for $T_{\rm BO}\!=\!600$ Adam steps at learning rate of 0.01.
Each BO step uses the target and all selected non-target cameras, with $(\lambda_{\rm tar},\lambda_{\rm non},\lambda_{\rm out})\!=\!(50,500,1)$.

\subsection{Main Results}
\label{sec:main-results}


\textbf{Recognition accuracy.}
We use \texttt{GPT-6 luna} with high reasoning effort to assess watermark recognition.
For the glyph \texttt{A}, \Cref{tab:exp-recognition} reports results across six settings.
The final state $\mathbf C$ achieves 91.00--100.00\% target accuracy, while reducing non-target recognition from the watermark-only state $\mathbf A$ to 0.00--3.75\% for Seen views and 0.00--3.13\% for Unseen views.
To distinguish watermark recognition from scene content, \Cref{tab:exp-target-recognition} compares watermarked target renderings with clean controls for 26 letters across ten scenes per letter.
Given a glyph-style reference, the evaluator correctly transcribes the intended letter in 89.62\% of watermarked images.
Clean controls receive no letter predictions, with 10.77\% Uncertain and 89.23\% None responses, supporting recognition attributable to the embedded watermark.

\begin{table}[t]
\centering
\caption{26-letter recognition outcomes (\%) for each condition.}
\label{tab:exp-target-recognition}
\vspace{-3mm}

\footnotesize
\setlength{\tabcolsep}{2pt}
\setlength{\aboverulesep}{0.3ex}
\setlength{\belowrulesep}{0.3ex}
\renewcommand{\arraystretch}{1.0}

\begin{tabular*}{\columnwidth}{@{\extracolsep{\fill}}lrrrr@{}}
\toprule
Scene & Correct & Other & Uncertain & None \\
\midrule
Watermarked & 89.62 & 9.23 & 0.38 & 0.77 \\
Original    & --    & --   & 10.77 & 89.23 \\
\bottomrule
\end{tabular*}
\vspace{-3mm}
\end{table}

\begin{table}[t]
\centering
\caption{Conditional non-target recognition.
At each offset, we retain $N$ cameras where the intended letter is recognized in $\mathbf A$ and evaluate $\mathbf C$ at those cameras.
Correct, Other, Uncertain, and None are the intended letter, a different letter, an ambiguous prediction, and no recognized letter, respectively.
Per-offset rates are $\frac{count}{N}$.
}
\label{tab:exp-offtarget-recognition}
\vspace{-3mm}

\footnotesize
\setlength{\tabcolsep}{1.5pt}
\setlength{\aboverulesep}{0.3ex}
\setlength{\belowrulesep}{0.3ex}
\renewcommand{\arraystretch}{1.05}

\resizebox{\columnwidth}{!}{%
\begin{tabular}{@{}lc|cccc|cccc@{}}
\toprule
\multirow{2}{*}{Offset}
& \multirow{2}{*}{\shortstack{$\mathbf A$ correct\\$N$}}
& \multicolumn{4}{c|}{$\mathbf C$: Outcome rates (\%)}
& \multicolumn{4}{c}{$\mathbf C$: Counts} \\
\cmidrule(lr){3-6}
\cmidrule(lr){7-10}
& & Correct & Other & Uncertain & None
& Correct & Other & Uncertain & None \\
\midrule

$0.05^\circ$ & 5,748
& 0.31 & 2.78 & 20.79 & 76.11
& 18 & 160 & 1,195 & 4,375 \\

$0.1^\circ$ & 4,169
& 0.07 & 1.89 & 12.59 & 85.44
& 3 & 79 & 525 & 3,562 \\

$0.25^\circ$ & 2,546
& 0.04 & 0.51 & 4.91 & 94.54
& 1 & 13 & 125 & 2,407 \\

$0.5^\circ$ & 2,108
& 0.00 & 0.14 & 2.61 & 97.25
& 0 & 3 & 55 & 2,050 \\

$1^\circ$ & 1,892
& 0.05 & 0.16 & 2.96 & 96.83
& 1 & 3 & 56 & 1,832 \\

$2^\circ$ & 990
& 0.20 & 0.30 & 3.03 & 96.46
& 2 & 3 & 30 & 955 \\

\midrule
\rowcolor{gray!8}
Avg. & 2,908.83
& 0.11 & 0.97 & 7.82 & 91.11
& 4.17 & 43.50 & 331.00 & 2,530.17 \\
\bottomrule
\end{tabular}%
}
\vspace{-5mm}
\end{table}

\begin{table*}[t]
\centering
\caption{Fidelity on feed-forward Gaussian scenes relative to clean renderings.
Full-image / local metrics report $\mathbf{B}\rightarrow \mathbf{C}$ / $\mathbf{A}\rightarrow \mathbf{C}$.
Local evaluation uses an $8\times8$ crop centered on the watermark projection.
Seen/Unseen indicate inclusion/exclusion from sidewall optimization.}
\label{tab:exp-local-ff}
\vspace{-3mm}

\footnotesize
\setlength{\tabcolsep}{2pt}
\setlength{\aboverulesep}{0.3ex}
\setlength{\belowrulesep}{0.3ex}
\renewcommand{\arraystretch}{1.0}

\begin{tabular*}{\textwidth}{@{\extracolsep{\fill}}ll*{8}{c}@{}}
\toprule
\multirow{2}{*}{Dataset}
& \multirow{2}{*}{View}
& \multicolumn{4}{c}{Cavity Splatting on MVSplat}
& \multicolumn{4}{c}{Cavity Splatting on IDESplat} \\
\cmidrule(lr){3-6}\cmidrule(lr){7-10}
&
& Full PSNR $\uparrow$ & Full SSIM $\uparrow$
& Local PSNR $\uparrow$ & Local SSIM $\uparrow$
& Full PSNR $\uparrow$ & Full SSIM $\uparrow$
& Local PSNR $\uparrow$ & Local SSIM $\uparrow$ \\
\midrule

\rowcolor{lightpurple}
\cellcolor{white}\multirow{3}{*}{RE10K}
& Target
& $35.68\!\rightarrow\!36.12$
& $.998\!\rightarrow\!.997$
& $5.67\!\rightarrow\!6.30$
& $.0727\!\rightarrow\!.0668$
& $35.56\!\rightarrow\!36.15$
& $.997\!\rightarrow\!.997$
& $5.54\!\rightarrow\!6.29$
& $.0551\!\rightarrow\!.0504$ \\

& Seen
& $23.20\!\rightarrow\!38.82$
& $.971\!\rightarrow\!.991$
& $9.03\!\rightarrow\!20.21$
& $.1763\!\rightarrow\!.6137$
& $23.17\!\rightarrow\!38.95$
& $.971\!\rightarrow\!.992$
& $8.73\!\rightarrow\!18.28$
& $.1732\!\rightarrow\!.5958$ \\

& Unseen
& $35.11\!\rightarrow\!36.25$
& $.997\!\rightarrow\!.997$
& $5.94\!\rightarrow\!7.78$
& $.0844\!\rightarrow\!.0802$
& $35.77\!\rightarrow\!36.63$
& $.997\!\rightarrow\!.997$
& $5.71\!\rightarrow\!8.00$
& $.0639\!\rightarrow\!.0789$ \\
\midrule

\rowcolor{lightpurple}
\cellcolor{white}\multirow{3}{*}{ACID}
& Target
& $35.91\!\rightarrow\!36.34$
& $.997\!\rightarrow\!.997$
& $5.90\!\rightarrow\!6.53$
& $.0482\!\rightarrow\!.0457$
& $36.07\!\rightarrow\!36.69$
& $.997\!\rightarrow\!.997$
& $6.09\!\rightarrow\!6.82$
& $.0517\!\rightarrow\!.0485$ \\

& Seen
& $24.77\!\rightarrow\!40.07$
& $.973\!\rightarrow\!.991$
& $9.28\!\rightarrow\!20.38$
& $.1578\!\rightarrow\!.5849$
& $24.82\!\rightarrow\!39.85$
& $.973\!\rightarrow\!.992$
& $9.02\!\rightarrow\!19.25$
& $.1609\!\rightarrow\!.5592$ \\

& Unseen
& $35.65\!\rightarrow\!36.70$
& $.997\!\rightarrow\!.997$
& $6.22\!\rightarrow\!8.15$
& $.0626\!\rightarrow\!.0641$
& $36.36\!\rightarrow\!37.41$
& $.997\!\rightarrow\!.997$
& $6.32\!\rightarrow\!8.64$
& $.0635\!\rightarrow\!.0724$ \\
\bottomrule
\end{tabular*}
\vspace{-5mm}
\end{table*}

\begin{table}[t]
\centering
\caption{Fidelity on 3DGS scenes from MIP and NS. States and metric conventions follow Table~\ref{tab:exp-local-ff}.}
\label{tab:exp-local-opt-alphabet}
\vspace{-3mm}

\footnotesize
\setlength{\tabcolsep}{2pt}
\setlength{\aboverulesep}{0.3ex}
\setlength{\belowrulesep}{0.3ex}
\renewcommand{\arraystretch}{1.0}

\resizebox{\columnwidth}{!}{%
\begin{tabular}{@{}ll*{4}{c}@{}}
\toprule
Dataset & View
& Full PSNR $\uparrow$ & Full SSIM $\uparrow$
& Local PSNR $\uparrow$ & Local SSIM $\uparrow$ \\
\midrule

\rowcolor{lightpurple}
\cellcolor{white}\multirow{3}{*}{MIP}
& Target
& $38.47\!\rightarrow\!39.15$
& $.997\!\rightarrow\!.997$
& $8.45\!\rightarrow\!9.33$
& $.1048\!\rightarrow\!.0807$ \\

& Seen
& $27.36\!\rightarrow\!40.48$
& $.980\!\rightarrow\!.993$
& $10.34\!\rightarrow\!19.77$
& $.1975\!\rightarrow\!.6518$ \\

& Unseen
& $41.35\!\rightarrow\!42.07$
& $.998\!\rightarrow\!.998$
& $8.56\!\rightarrow\!11.11$
& $.1112\!\rightarrow\!.0945$ \\
\midrule

\rowcolor{lightpurple}
\cellcolor{white}\multirow{3}{*}{NS}
& Target
& $36.72\!\rightarrow\!37.01$
& $.997\!\rightarrow\!.997$
& $6.72\!\rightarrow\!7.31$
& $.0842\!\rightarrow\!.0898$ \\

& Seen
& $23.78\!\rightarrow\!35.89$
& $.968\!\rightarrow\!.987$
& $9.10\!\rightarrow\!17.19$
& $.2190\!\rightarrow\!.4178$ \\

& Unseen
& $37.40\!\rightarrow\!36.93$
& $.997\!\rightarrow\!.997$
& $6.78\!\rightarrow\!8.44$
& $.0851\!\rightarrow\!.0849$ \\
\bottomrule
\end{tabular}%
}
\vspace{-3mm}
\end{table}

\begin{table}[t]
\centering
\caption{Fidelity across 26 letters on 100 RealEstate10K scenes with MVSplat. States and metric conventions follow Table~\ref{tab:exp-local-ff}.}
\label{tab:exp-alphabet-fidelity}
\vspace{-3mm}

\footnotesize
\setlength{\tabcolsep}{2pt}
\setlength{\aboverulesep}{0.3ex}
\setlength{\belowrulesep}{0.3ex}
\renewcommand{\arraystretch}{1.0}

\resizebox{\columnwidth}{!}{%
\begin{tabular}{@{}l*{4}{c}@{}}
\toprule
View
& Full PSNR $\uparrow$ & Full SSIM $\uparrow$
& Local PSNR $\uparrow$ & Local SSIM $\uparrow$ \\
\midrule

\rowcolor{lightpurple}
Target
& $36.16\!\rightarrow\!36.70$
& $.997\!\rightarrow\!.997$
& $6.25\!\rightarrow\!6.90$
& $.0843\!\rightarrow\!.0767$ \\

Seen
& $23.93\!\rightarrow\!38.81$
& $.973\!\rightarrow\!.992$
& $8.82\!\rightarrow\!19.84$
& $.1810\!\rightarrow\!.6097$ \\

Unseen
& $35.71\!\rightarrow\!36.77$
& $.997\!\rightarrow\!.997$
& $6.41\!\rightarrow\!8.40$
& $.0951\!\rightarrow\!.0915$ \\
\bottomrule
\end{tabular}%
}
\vspace{-4mm}
\end{table}

\vspace{1mm}\noindent\textbf{Conditional non-target recognition test.}
We evaluate non-target recognition under small camera rotations, finding that optimized sidewalls strongly suppress letter recognition even at a $0.05^\circ$ offset, conditional on correct recognition in the watermark-only state. We further evaluate 26 letters on a common set of 50 scenes across all 26 letters using MLLM-based tests under bidirectional yaw, pitch, and diagonal rotations at offsets $\{0.05,0.1,0.25,0.5,1,2\}^\circ$.
For this recognition scan, diagonal rotations are normalized
to the same total SO(3) angle as yaw and pitch.
\Cref{tab:exp-offtarget-recognition} reports mean outcome rates of 91.11\% None, 7.82\% Uncertain, and 0.97\% Other across offsets.
Correct recognition averages only 0.11\% and is just 0.31\% at the smallest $0.05^\circ$ offset, demonstrating strong non-target suppression even under very small camera rotations.
\vspace{1mm}\noindent\textbf{Full-image and local-area fidelity.}
We evaluate fidelity relative to clean renderings, finding substantial full-image and local-area improvements at Seen views and consistent local PSNR gains at Unseen views.
\Cref{tab:exp-local-ff,tab:exp-local-opt-alphabet} report full-image fidelity from $\mathbf{B}$ to $\mathbf{C}$ and local-area fidelity from $\mathbf{A}$ to $\mathbf{C}$.
Across the six model--dataset settings, Seen-view PSNR increases by 12.11--15.78 for full images and 8.09--11.18 within the $8\times8$ local area, with higher SSIM in both regions.
These gains demonstrate reduced artifacts from the initialized sidewalls and improved local appearance over the watermark-only state.
At Unseen views, full-image PSNR improves in five settings and local PSNR in all six, although local SSIM changes are mixed.
The 26-letter evaluation in \Cref{tab:exp-alphabet-fidelity} confirms improvements in Seen-view PSNR and SSIM at both scales, extending these benefits across glyphs.
Target full-image PSNR also improves across all evaluations.
Together, these results support effective sidewall blending and scene appearance preservation across reconstruction methods and watermark content.

\vspace{1mm}
\noindent\textbf{Projection incidence \& fidelity.}
We assess projection incidence in original dataset views and fidelity where projections remain (\Cref{tab:exp-projection}).
Native-frame incidence is only 0.54--1.66\% for feed-forward models, 4.74\% for MIP, and 1.25\% for NS, leaving most views without added-structure projections.
Fidelity is measured only on native-frame views retaining projections at $\mathbf C$, avoiding error dilution by empty-projection views.
LPIPS remains at or below 0.018 across all settings, indicating low average perceptual distortion even where the added structure projects.

\begin{table}[t]
\centering
\caption{Projection incidence and fidelity at $\mathbf C$.
Native/256-crop entries report count / rate (\%) of cameras with nonempty watermark or sidewall support.
Fidelity compares native-frame renderings retaining such projections against clean renderings.}
\label{tab:exp-projection}
\label{tab:exp-remaining}
\vspace{-3mm}

\fontsize{8}{9}\selectfont
\setlength{\tabcolsep}{3pt}
\setlength{\aboverulesep}{0.3ex}
\setlength{\belowrulesep}{0.3ex}
\renewcommand{\arraystretch}{1.0}

\resizebox{\columnwidth}{!}{%
\begin{tabular}{@{}lrrrrr@{}}
\toprule
Model/Dataset
& Cameras & Native & 256 crop
& PSNR $\uparrow$ & LPIPS $\downarrow$ \\
\midrule
MVSplat/RE10K
& 960,117 & 7,201/0.75 & 4,226/0.44 & 33.54 & 0.010 \\
MVSplat/ACID
& 455,878 & 2,477/0.54 & 1,622/0.36 & 34.51 & 0.006 \\
IDESplat/RE10K
& 148,890 & 841/0.57 & 744/0.50 & 26.62 & 0.018 \\
IDESplat/ACID
& 455,878 & 7,575/1.66 & 6,468/1.42 & 32.30 & 0.008 \\
3DGS/MIP
& 1,940 & 92/4.74 & 69/3.56 & 74.99 & 0.000 \\
3DGS/NS
& 2,400 & 30/1.25 & 30/1.25 & 42.97 & 0.003 \\
\bottomrule
\end{tabular}%
}
\vspace{-3mm}
\end{table}


\begin{table}[t]
\centering
\caption{Recognition retention among the 96 of 100 images where \texttt{A} is initially recognized. Retention measures the percentage remaining recognizable after perturbation. Perb. denotes perturbation.
For merging, strength is Gaussian-count reduction. }
\label{tab:attack-retention}
\vspace{-3mm}
\footnotesize
\setlength{\tabcolsep}{2pt}
\renewcommand{\arraystretch}{1.08}
\resizebox{\columnwidth}{!}{%
\begin{tabular}{@{}lrrrr@{}}
\toprule
Removal methods
& \multicolumn{4}{c}{Strength / Retention (\%) $\uparrow$} \\
\midrule
Position Perb.
& $5\%/97.92$ & $10\%/97.92$ & $20\%/94.79$ & $40\%/93.75$ \\
Scale Perb.
& $0.02/98.96$ & $0.05/100.00$ & $0.10/96.88$ & $0.20/97.92$ \\
Gaussian merging
& $5\%/96.88$ & $10\%/97.92$ & $20\%/100.00$ & $40\%/95.83$ \\
Outlier removal
& $0.1\%/98.96$ & $0.5\%/97.92$ & $1\%/100.00$ & $2\%/96.88$ \\
Random pruning
& $1\%/96.88$ & $5\%/96.88$ & $10\%/94.79$ & $20\%/79.17$ \\
\bottomrule
\end{tabular}
}
\vspace{-4mm}
\end{table}

\begin{figure*}[t]
\centering
\includegraphics[width=0.86\textwidth]{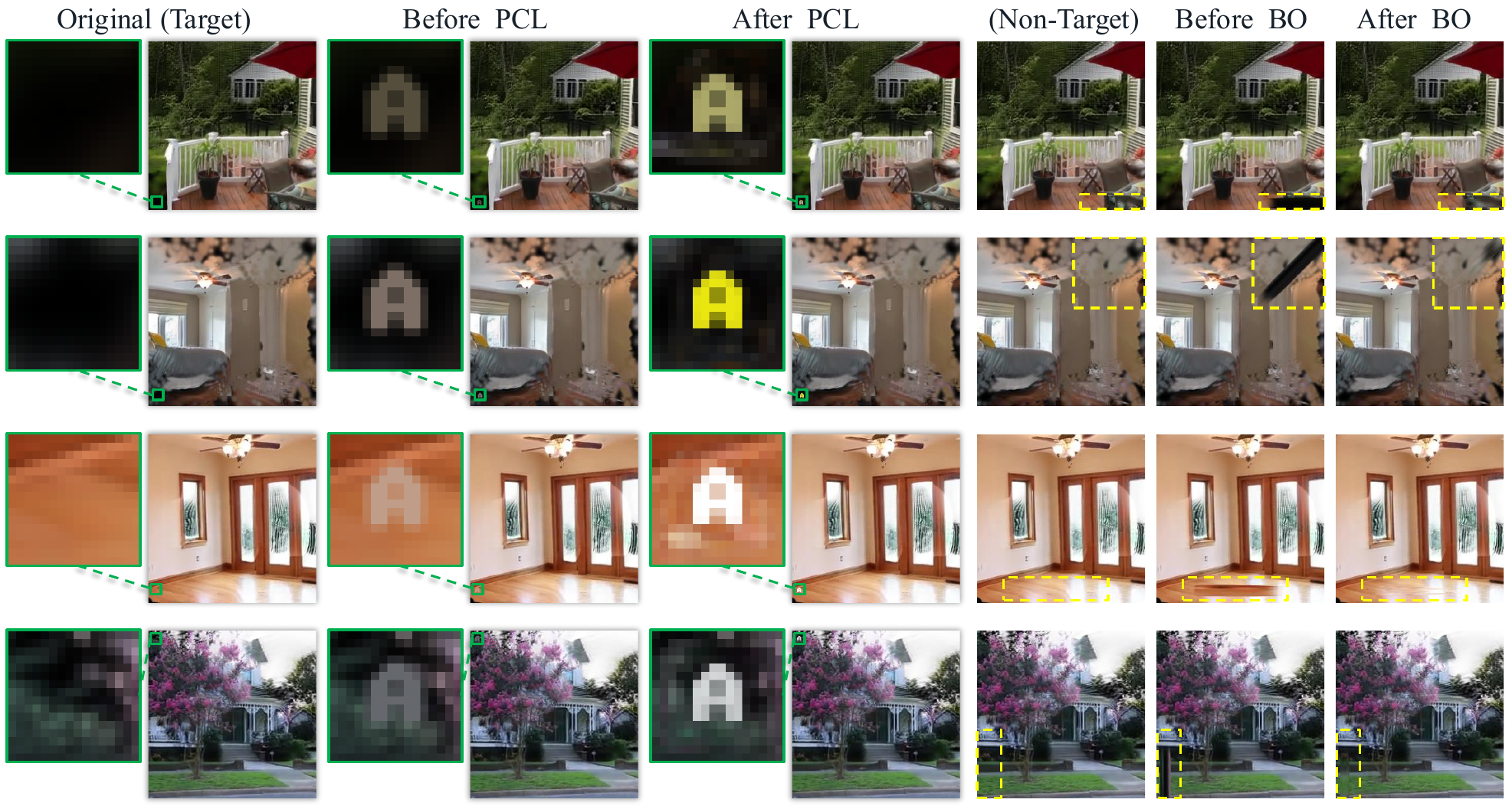}
\vspace{-4mm}
\caption{Target and non-target view appearance before and after PCL and BO. Each row shows the original scene, the watermark initialized from the global mean color, and the result after color optimization. Green boxes locate the identical regions enlarged beside each rendering.}
\label{fig:exp-pcl-visualization}
\vspace{-4mm}
\end{figure*}






\vspace{1mm}\noindent\textbf{Robustness against watermark removals.}
We evaluate recognition retention under geometric perturbations and Gaussian reduction operations, finding strong robustness overall, with reduced retention under extensive random pruning.
Table~\ref{tab:attack-retention} reports 93.75\%--100\% retention
under position/scale perturbations, merging, and outlier removal,
including 95.83\% after a 40\% reduction by merging.
Random pruning retains 94.79\% and 79.17\% at 10\% and 20\%
removal, respectively.

\vspace{1mm}\noindent\textbf{Visualization Results.}
\Cref{fig:exp-pcl-visualization} compares target renderings before and after PCL and BO with the same scene and placement. The successful cases illustrate different solutions: a light letter against dark vegetation, a dark letter on a light wall, and a chromatic letter on a warm background. The examples illustrate changes in luminance and color separation across different backgrounds.

\subsection{Ablation Study and Discussion}
\label{sec:ablation}

\begin{table}[t]
\centering
\caption{Cavity-layer ablation.
Target accuracy uses 100 images.
Retention uses the fixed 96 baseline-recognized scenes, with baseline judgments reused for the default 96/96 entry.
Rotation accuracy uses 720 views, while strict hiding uses 282 readable watermark-only views and excludes wrong-letter responses.
Metrics are percentages except Other (wrong-letter count).}
\label{tab:ablation-cavity-layers}
\vspace{-3mm}

\footnotesize
\setlength{\tabcolsep}{2pt}
\setlength{\aboverulesep}{0.3ex}
\setlength{\belowrulesep}{0.3ex}
\renewcommand{\arraystretch}{1.0}

\begin{tabular*}{\columnwidth}{@{\extracolsep{\fill}}l*{5}{c}@{}}
\toprule
Layers
& Target acc.$\uparrow$
& Retention$\uparrow$
& Rot. acc.$\downarrow$
& Strict hiding$\uparrow$
& Other \\
\midrule
0 & 96.00 & 97.92  & 39.17 & 0.00  & 0 \\
1 & 96.00 & 97.92  & 2.08  & 93.26 & 4 \\
\rowcolor{gray!8}
2 & 96.00 & 100.00 & 0.42  & 97.87 & 3 \\
3 & 94.00 & 95.83  & 0.14  & 98.94 & 2 \\
\bottomrule
\end{tabular*}
\vspace{-4mm}
\end{table}

\noindent\textbf{Cavity layers.}
We test zero to three layers on 100 MVSplat/RE10K scenes (Table~\ref{tab:ablation-cavity-layers}).
Strict hiding is assessed on 282 readable watermark-only views
among 720 rotated views from 20 scenes.
It rises from 0\% without a cavity to 93.26\% with one layer
and 97.87\% with two, while target accuracy stays at 96\%.
Three layers raise hiding to 98.94\% but reduce target accuracy
to 94\%, supporting two layers as a tradeoff between suppression
and target readability.


\vspace{1mm}\noindent\textbf{PCL loss weights.}
Table~\ref{tab:color-loss} varies the relative weights of WCAG contrast and CIEDE2000.
Combining both terms yields 96--97\% target accuracy and 97.92--100\% retention, compared with 91--92\% accuracy using either term alone and 54\% without color optimization.
These results demonstrate the benefit of jointly optimizing contrast and color separation for reliable watermark recognition.

\begin{table}[t]
\centering
\caption{PCL loss-weight ablation study (WCAG contrast:CIEDE2000).
The default is $1{:}1$.
Acc. uses 100 targets, and retention uses the fixed 96 baseline-recognized targets.}
\label{tab:color-loss}
\vspace{-3mm}
\small
\setlength{\tabcolsep}{2pt}
\setlength{\aboverulesep}{0.3ex}
\setlength{\belowrulesep}{0.3ex}
\renewcommand{\arraystretch}{1.0}

\begin{tabular*}{\columnwidth}{@{\extracolsep{\fill}}l*{6}{c}@{}}
\toprule
Loss ratio
& $1{:}0$ & $1{:}0.5$
& \cellcolor{gray!8}$1{:}1$
& $1{:}2$ & $0{:}1$ & No color opt. \\
\midrule
Acc. (\%) $\uparrow$
& 91 & 96 & \cellcolor{gray!8}96 & 97 & 92 & 54 \\
Retention (\%) $\uparrow$
& 93.75 & 97.92 & \cellcolor{gray!8}100.00
& 98.96 & 93.75 & 55.21 \\
\bottomrule
\end{tabular*}
\vspace{-4mm}
\end{table}

\vspace{1mm}\noindent\textbf{More ablation studies}
on $q_{\mathrm{out}}$, cavity geometry,
the lateral scale of watermark Gaussians, training-view selection
and count, other loss weights, and other factors are in the
appendix. The results show tradeoffs among target-view
readability, off-target suppression, and rendering fidelity.

\vspace{1mm}\noindent\textbf{Ownership verification.}
Our hypothesis tests provide strong statistical support for ownership verification based on high target-view watermark recognition and effective suppression of correct non-target recognition.
Detailed hypotheses, procedures, and results are in the appendix.

\section{Conclusion}
\label{sec:conclusion}
We presented Cavity Splatting for view-selective visible
watermarking of already reconstructed Gaussian scenes.
Boundary-aware placement and ray-aligned sidewalls control
where the watermark is exposed, PCL improves its contrast,
and BO adapts the added structure while keeping the original
scene fixed, reducing watermark leakage and restoring
local-view fidelity. Experiments show low watermark exposure
and limited appearance changes in the scene's original
observation views, while maintaining target-view glyph
legibility and markedly reducing correct watermark recognition
in nearby non-target views. Results across multiple datasets,
two feed-forward reconstruction methods, and one
optimization-based method support the method's applicability
across different datasets and reconstruction approaches, as
well as its view selectivity under the evaluated camera
protocols. However, watermark concealment remains less
effective under camera translations than under rotational
perturbations. Future work will explore cavity geometries and
optimization strategies to improve concealment under camera
translations.


\newpage
\section*{LLM Usage Statement}
We used GPT-6 Astra to help revise the code for our LLM-based watermark recognition evaluation pipeline and to edit the manuscript for clarity and readability.
For code development, it assisted in implementing parts of the experimental scripts, as well as scripts for generating tables and figures.
 We used GPT-6 Luna to assess watermark recognition and Kimi3 to assess glyph readability in the supplementary matched-view experiment. The authors reviewed the code and manuscript revisions and take full responsibility for the evaluation procedure and all reported results.

{
    \small
    \bibliographystyle{ieeenat_fullname}
    \bibliography{main}
}

\clearpage
\setcounter{page}{1}
\maketitlesupplementary

\section{Additional Details of Methodology}
\label{sec:additional-methodology}

We expand the camera conventions, placement search, Gaussian construction,
and optimization procedure in Sec.~\ref{sec:method}.
The notation follows the main paper. Equations already given there are
referenced below. Evaluation definitions distinguish geometric projection,
image fidelity, color preservation, and letter recognition.

\subsection{Camera Coordinates and Rendering}
\label{sec:ew-supp-cameras}
\label{sec:ew-supp-overview}

\subsubsection{Coordinate and pixel conventions}
For the camera-to-world pose $P_k=(K_k,R_k,t_k)$ used in
Eq.~\eqref{eq:3dgs_encoder_decoder}, the homogeneous transform and
world-to-camera conversion are
\begin{equation}
 T_k=\begin{bmatrix}R_k&t_k\\0^\top&1\end{bmatrix},
 \qquad x_{\rm cam}=R_k^\top(x_{\rm world}-t_k).
 \label{eq:ew-supp-camera-coordinates}
\end{equation}
Writing $K_k$ for pixel-space intrinsics, normalized intrinsics satisfy
\begin{equation}
 S_{HW}=\operatorname{diag}(W^{-1},H^{-1},1),
 \qquad \bar K_k=S_{HW}K_k.
 \label{eq:ew-supp-normalized-intrinsics}
\end{equation}
The zero-based pixel-center convention is the one in
Eq.~\eqref{eq:ew-main-watermark-backprojection}.
Multiplying the homogeneous pixel coordinate by $S_{HW}$ and using
$\bar K_k$ gives the same ray, since $\bar K_k^{-1}S_{HW}=K_k^{-1}$.
The ray's third component is one: its depth parameter is optical-axis
depth, not Euclidean distance along a unit ray.

\subsubsection{Projection and compositing}
For camera coordinates $x=(X,Y,Z)^\top$, the pixel projection and its
Jacobian are
\begin{equation}
 \begin{aligned}
 \pi_K(x)&=\begin{bmatrix}f_xX/Z+c_x\\f_yY/Z+c_y\end{bmatrix},\\
 J(x)&=\begin{bmatrix}
 f_x/Z&0&-f_xX/Z^2\\0&f_y/Z&-f_yY/Z^2
 \end{bmatrix}.
 \end{aligned}
 \label{eq:ew-supp-projection-jacobian}
\end{equation}
Ignoring discrete rasterizer truncation, the projected covariance of
primitive $j$ is approximated by
\begin{equation}
 \Sigma_{jk}^{2D}
 =J_{jk}R_k^\top\Sigma_jR_kJ_{jk}^\top+\Sigma_{\rm filter},
 \label{eq:ew-supp-projected-covariance}
\end{equation}
where $J_{jk}$ is evaluated at the camera-space primitive center and
$\Sigma_{\rm filter}$ denotes the rasterizer's filtering contribution.
Actual projection supports are computed with the rendering implementation,
including its clipping and threshold rules.

For primitives sorted front to back at camera $k$, let $a_{jk}(p)$ be
primitive $j$'s effective per-pixel opacity, $d_{jk}$ its viewing direction,
and $c_{\rm bg}$ the background color. Alpha compositing gives
\begin{equation}
 \begin{aligned}
 I_k(p)&=\sum_j T_{jk}(p)a_{jk}(p)\rho_j(d_{jk})
        +T_{{\rm end},k}(p)c_{\rm bg},\\
 T_{jk}(p)&=\prod_{i<j}(1-a_{ik}(p)).
 \end{aligned}
 \label{eq:ew-supp-alpha-compositing}
\end{equation}
Here $T_{{\rm end},k}$ is the residual transmittance after all contributing
primitives. The SH color model is already given in
Eq.~\eqref{eq:ew-main-sh-color}.
Freezing watermark parameters does not freeze its final pixel contribution:
sidewall motion and opacity change transmittance and compositing weights.

\subsection{Crop-Consistent Camera Intrinsics}
\label{sec:ew-supp-intrinsics}

\subsubsection{Dataset camera conversion}
MVSplat uses the saved per-frame normalized intrinsics and camera poses.
For MIP, COLMAP \texttt{PINHOLE} supplies separate focal lengths, while
\texttt{SIMPLE\_PINHOLE} uses one focal length for both axes.
For NS, the native horizontal field of view gives
\begin{equation}
 f_x^{\rm native}=\frac{W_n}{2\tan(\mathrm{FOV}_x/2)},
 \qquad f_y^{\rm native}=f_x^{\rm native}.
 \label{eq:ew-supp-ns-focal}
\end{equation}
The NS adapter flips the second and third columns of the Blender rotation
to obtain the OpenCV camera convention. It merges train/test cameras and
removes duplicate calibrated poses. MIP registered cameras are sorted by
image name, with every eighth camera marked as a test camera.

Let $(W_n,H_n)$ be the native image size and $R_c=256$ the square crop size.
Resizing to cover the crop uses
\begin{equation}
 \begin{aligned}
 s_{\rm resize}&=\max(R_c/W_n,R_c/H_n),\\
 W'&=\operatorname{round}(s_{\rm resize}W_n),\quad
 H'=\operatorname{round}(s_{\rm resize}H_n).
 \end{aligned}
 \label{eq:ew-supp-crop-resize}
\end{equation}
The crop-adjusted normalized focal lengths are
\begin{equation}
 \bar f_x^{\rm crop}=\bar f_x^{\rm native}\frac{W'}{R_c},
 \qquad
 \bar f_y^{\rm crop}=\bar f_y^{\rm native}\frac{H'}{R_c}.
 \label{eq:ew-supp-crop-focal}
\end{equation}
For the NS/MIP adapters, the principal point also includes the integer
center-crop offset:
\begin{equation}
 \begin{aligned}
 \bar c_x^{\rm crop}
 &=\frac{W'\bar c_x^{\rm native}-\lfloor(W'-R_c)/2\rfloor}{R_c},\\
 \bar c_y^{\rm crop}
 &=\frac{H'\bar c_y^{\rm native}-\lfloor(H'-R_c)/2\rfloor}{R_c}.
 \end{aligned}
 \label{eq:ew-supp-crop-principal-point}
\end{equation}
Thus, a cropped dataset camera need not have a principal point
at exactly $(0.5,0.5)$. MVSplat follows its official crop shim, checks that the input
principal point is centered, and retains this centered value.

\subsubsection{Target versus evaluation intrinsics}
The target and local candidates use
Eq.~\eqref{eq:ew-main-target-intrinsics} with $\kappa=2.5$ and a centered
principal point. The two focal-length maxima are computed separately over
the current scene's crop-adjusted cameras and can come from different
frames. For the centered target camera,
\begin{equation}
 \mathrm{FOV}_{a,\star}
 =2\arctan\!\left(\frac{1}{2\bar f_a^\star}\right),
 \qquad a\in\{x,y\}.
 \label{eq:ew-supp-target-fov}
\end{equation}
Changing output resolution while keeping normalized intrinsics fixed
preserves the field of view. Center cropping and the target focal
multiplier instead change it. Original-camera evaluation uses each
camera's own native or crop-adjusted intrinsics, without the target
multiplier. Its clean reference is a same-camera rendering of the frozen
Gaussian scene, rather than the dataset photograph.

\subsection{Finite-Budget Boundary-Aware Placement}
\label{sec:ew-supp-placement}

\subsubsection{Pose search and corner placement}
The initial pose $(R_0,t_0)$ is the first target camera in the MVSplat test
configuration, or the first test camera in the NS/MIP camera list.
Let $Z_i$ and $(u_i^{\rm norm},v_i^{\rm norm})$ be the depth and normalized
projection of original Gaussian center $i$. The reference scene scale is
\begin{equation}
 \begin{aligned}
 \mathcal I_D=\{i:&z_{\rm near}<Z_i<z_{\rm far},\ \alpha_i>0.1,\\
                  &0.2<u_i^{\rm norm}<0.8,\
                   0.2<v_i^{\rm norm}<0.8\},\\
 D&=\operatorname{median}_{i\in\mathcal I_D}Z_i.
 \end{aligned}
 \label{eq:ew-supp-reference-depth}
\end{equation}
This is a statistic of Gaussian centers, not a recovered visible-surface
depth. An empty set is treated as an error. MVSplat computes it before
replacing the initial crop intrinsics. For both feed-forward and optimization-based reconstructions,
we compute $D$ using the starting camera's crop-adjusted
intrinsics before applying the target focal enlargement. Candidate translations are parameterized as
\begin{equation}
 t(a,b,d)=t_0+DR_0[a,b,-d]^\top.
 \label{eq:ew-supp-camera-search}
\end{equation}
The search first retreats along the optical axis in steps of $0.025D$ to
find low coverage, then explores deeper positions, corner-aligning
rotations, and lateral translations. The configured retreat range reaches
$d=4$, with lateral offsets $a,b\in[-0.22,0.22]$.

The low-coverage mask and strict criteria are defined in
Eqs.~\eqref{eq:ew-main-low-coverage} and
\eqref{eq:ew-main-placement-criteria}.
We use $\epsilon_\alpha=0.01$, four $32\times32$ corner windows, and
$n_{\min}=h_{\min}=4$.
Within the selected corner, the $8\times8$ glyph origin is searched over
integer offsets from 8 to 16 along each axis, maximizing the stroke
low-coverage count $N_P(M_\xi)$.
Ties prefer a smaller alpha sum in the surrounding $14\times14$ square
excluding the inner glyph box. The strict test requires zero low-coverage
pixels outside the selected corner. It does not require the entire glyph
box or cavity projection to lie in a low-coverage region.

\subsubsection{Thresholded clearance and frustum exclusion}
The clearance in Eq.~\eqref{eq:ew-main-clearance} uses a density threshold
$\tau=0.01$. For each active scene Gaussian, a sphere enclosing its
threshold ellipsoid has radius
\begin{equation}
 r_i=\sqrt{2\lambda_{\max}(\Sigma_i)\log(\alpha_i/\tau)},
 \qquad \alpha_i\geq\tau.
 \label{eq:ew-supp-threshold-radius}
\end{equation}
For the target ray $q_B$ through the glyph-box center, the template's
corridor sphere is
\begin{equation}
 \begin{aligned}
 b&=t_\star+0.225R_\star q_B,\\
 r_C&=0.225\|q_B\|_2+0.12\\
    &\quad+\frac{0.45\cdot8}{256}
       \sqrt{(\bar f_x^\star)^{-2}+(\bar f_y^\star)^{-2}}.
 \end{aligned}
 \label{eq:ew-supp-corridor-sphere}
\end{equation}
The factor $0.225$ is half the watermark depth, not the midpoint
$0.33$ of the cavity front and back. The 0.12 term is a fixed safety inflation factor determined through calculation. These thresholded bounding spheres do not truncate Gaussian
tails or prove zero-density separation.

Reference cameras are selected at approximately uniform frame-index
intervals, using at most 32 crop-adjusted cameras without focal
enlargement. Let $(n_{kh},\beta_{kh})$ denote their inward-facing frustum
planes with unit normals. The exclusion fraction is
\begin{equation}
 q_{\rm out}=\frac{1}{|\mathcal P_{\rm ref}|}
 \sum_{P_k\in\mathcal P_{\rm ref}}
 \mathbf 1\!\left[
 \min_h(n_{kh}^{\top}b+\beta_{kh})<-r_C\right].
 \label{eq:ew-supp-frustum-exclusion}
\end{equation}
The planes comprise the forward plane and four side planes. Exclusion
does not rely on a far plane. We set $q_{\min}=0.95$ in
Eq.~\eqref{eq:ew-main-placement-criteria}, requiring at least 31 exclusions
when 32 reference cameras are available. This tests the crop frusta and
is distinct from occlusion by the original scene.

\subsubsection{Search budget and fallback}
The search returns the first strict candidate within its 120-second
budget. Otherwise, it first ranks candidates with no low-coverage pixels
outside the chosen corner. Writing
\begin{equation}
 d_{\rm near}(P)=\min_i\|t_P-\mu_i\|_2,
 \qquad H(P)=N_P(M_{\xi(P)}),
 \label{eq:ew-supp-fallback-distance}
\end{equation}
this ranking lexicographically maximizes $(d_{\rm near},H)$.
If no such candidate exists, successfully rendered candidates are ranked
by $(-N_P(\Omega\setminus Q),d_{\rm near},H)$.
Here $d_{\rm near}$ uses all original Gaussian centers and measures camera
separation, whereas the main-paper clearance uses active Gaussians and the
corridor sphere. Neither clearance nor $q_{\rm out}$ is a hard fallback
constraint. A fallback continues to optimization with its relaxed status
recorded. It is not relabeled as a strict placement success.

\subsection{Watermark and Sidewall Geometry}
\label{sec:ew-supp-initialization}

\subsubsection{Watermark initialization}
The glyph is specified by an $8\times8$ IBM bitmap, with one Gaussian per active pixel. Thus, $N_W=|M|$ depends on the glyph. The letter \texttt{A} used here contains 28 active pixels. We place the watermark at target-camera depth $z_W=0.45$ using Eq.~\eqref{eq:ew-main-watermark-backprojection}, including its half-pixel offset for integer pixel indices. The lateral scales follow Eq.~\eqref{eq:ew-main-watermark-scales}
with $\sigma_p=0.33$ pixels. To minimize the impact on
non-target views, we set the normal scale to
\begin{equation}
    s_z=0.01\min(s_x,s_y).
    \label{eq:ew-supp-watermark-normal-scale}
\end{equation}
We initialize watermark opacity to $0.99$ and use only the DC color coefficient under the SH convention of Eq.~\eqref{eq:ew-main-sh-color}. All higher-order coefficients are zero, and $\psi_0=0.2820947918$. PCL updates each Gaussian's RGB color independently while preserving its center, covariance, and opacity. After PCL, the entire watermark is frozen. These depths are absolute target-camera coordinates, so preserving the target image footprint through a joint change in depth and scale would still change parallax and occlusion at other cameras.

\subsubsection{Continuous aperture and depth sampling}
The cavity has two layers, each comprising four sidewalls with $n_{\mathrm{edge}}=n_{\mathrm{depth}}=8$, giving 512 Gaussians. Both ends remain open. Its eight depth samples include both endpoints:
\begin{equation}
    \begin{aligned}
        z_k&=z_{\mathrm{front}}
        +\frac{k}{n_{\mathrm{depth}}-1}
        (z_{\mathrm{back}}-z_{\mathrm{front}}),\\
        k&=0,\ldots,7,
        \qquad
        (z_{\mathrm{front}},z_{\mathrm{back}})=(0.21,0.45).
    \end{aligned}
    \label{eq:ew-supp-sidewall-depth-grid}
\end{equation}
Let $(x_W,y_W)$ denote the upper-left integer coordinate of the glyph box. For layer $\ell\in\{0,1\}$, the continuous aperture rectangle is
\begin{equation}
    \begin{aligned}
        m_\ell&=2.5+\ell/6,\\
        (x_0,y_0)&=(x_W-m_\ell,\ y_W-m_\ell),\\
        (x_1,y_1)&=(x_W+8+m_\ell,\ y_W+8+m_\ell).
    \end{aligned}
    \label{eq:ew-supp-aperture-rectangle}
\end{equation}
We sample each of its four edges by linear interpolation at $t_e=(e+\tfrac12)/8$, for $e=0,\ldots,7$. These are continuous image coordinates, so their backprojection uses the convention of Eq.~\eqref{eq:ew-main-watermark-backprojection} without adding another half-pixel offset. Denote the resulting unnormalized camera ray of sidewall primitive $j$ by $q_j$, whose third component is one. At its assigned depth $z_k$, the initial target-camera center $x_j^0$ is $z_k$ times this ray.

\subsubsection{Local frames and initial scales}
We construct each fixed local frame in target-camera coordinates. The ray axis is $\hat r_j=q_j/\|q_j\|_2$. For a left or right wall, $\hat n_j$ is the normalized version of $(1,0,-q_{jx})^\top$. For a top or bottom wall, it is the normalized version of $(0,1,-q_{jy})^\top$. The remaining axis and frame are
\begin{equation}
    \hat e_j
    =\frac{\hat r_j\times\hat n_j}
    {\|\hat r_j\times\hat n_j\|_2},
    \qquad
    F_j=[\hat e_j\ \hat r_j\ \hat n_j].
    \label{eq:ew-supp-sidewall-frame}
\end{equation}
Let $J_j$ be the pixel-space projection Jacobian in Eq.~\eqref{eq:ew-supp-projection-jacobian}, evaluated at $x_j^0$. Write $a=x$ for left/right walls and $a=y$ for top/bottom walls. The initial standard deviations along the edge, ray, and normal axes are determined by
\begin{equation}
    \begin{aligned}
        \widetilde s_{j,e}
        &=\frac{[(8+2m_\ell)/n_{\mathrm{edge}}]\,
        \beta_{\mathrm{wall}}}{\|J_j\hat e_j\|_2},\\
        \widetilde s_{j,r}
        &=\frac{z_{\mathrm{back}}-z_{\mathrm{front}}}
        {n_{\mathrm{depth}}-1}\,
        \|q_j\|_2\beta_{\mathrm{wall}},\\
        \widetilde s_{j,n}
        &=\frac{t_{\mathrm{px}}}{|(J_j\hat n_j)_a|},
    \end{aligned}
    \label{eq:ew-supp-sidewall-base-scales}
\end{equation}
where $\beta_{\mathrm{wall}}=0.65$ and $t_{\mathrm{px}}=1/12$ pixel. We further shrink only the normal axis by $\gamma_{\mathrm{normal}}=0.25$:
\begin{equation}
    s_j^0
    =\bigl(\widetilde s_{j,e},\ \widetilde s_{j,r},\
    \gamma_{\mathrm{normal}}\widetilde s_{j,n}\bigr)^\top.
    \label{eq:ew-supp-sidewall-initial-scales}
\end{equation}
Consequently, the first-order projected normal standard deviation along the corresponding image axis is $\gamma_{\mathrm{normal}}t_{\mathrm{px}}=1/48$ pixel. This quantity does not specify the final rasterized support width, which also depends on filtering and opacity thresholds.

At the initial center, perspective projection satisfies
\begin{equation}
    J_jq_j=0.
    \label{eq:ew-supp-ray-null-direction}
\end{equation}
Extending a primitive along its initial target ray therefore does not increase its transverse footprint to first order. This property generally does not hold at other cameras. Moreover, $F_j$ stays fixed while centers can move, so the ray axis need not remain aligned with the direction from the target camera to the current center.

\subsubsection{Bounded geometry and appearance}
\label{sec:ew-supp-parameter-bounds}
The center variable $x_j$ is expressed in target-camera coordinates, while $F_j$ specifies the covariance axes. We derive its displacement bounds from the initial ellipsoid:
\begin{equation}
    \begin{aligned}
        b_j^{\mathrm{raw}}
        &=\beta_{\mathrm{pos}}|F_j|s_j^0,
        \qquad \beta_{\mathrm{pos}}=0.35,\\
        b_{j,x}&=b_{j,x}^{\mathrm{raw}},
        \qquad b_{j,y}=b_{j,y}^{\mathrm{raw}},\\
        b_{j,z}
        &=\min\bigl\{b_{j,z}^{\mathrm{raw}},\
        x_{j,z}^0-z_{\mathrm{front}},\\
        &\hspace{12mm}z_{\mathrm{back}}-x_{j,z}^0\bigr\}.
    \end{aligned}
    \label{eq:ew-supp-sidewall-displacement-bounds}
\end{equation}
Here, $|F_j|$ is the elementwise absolute value. For the frontmost and rearmost depth layers, we set $b_{j,z}=0$ while allowing transverse motion. These bounds define fixed axis-aligned regions around the initial centers. They do not imply disjoint solid cells.

Using unconstrained variables $u_j,v_j,a_j,g_j$ and the sigmoid function $\sigma$, the bounded parameters are
\begin{equation}
    \begin{aligned}
        x_j&=x_j^0+b_j\odot\tanh(u_j),\\
        s_j^{\mathcal C}
        &=s_j^0\odot\exp\!\left[
        \log0.75+\sigma(v_j)\log(1.5/0.75)\right],\\
        \alpha_j^{\mathcal C}
        &=0.01+0.98\,\sigma(a_j),\\
        \bar c_j&=\sigma(g_j).
    \end{aligned}
    \label{eq:ew-supp-bounded-sidewall-parameters}
\end{equation}
The nonlinearities and products act elementwise. Scale multipliers lie between $0.75$ and $1.5$, and opacities lie between $0.01$ and $0.99$. The world-space center and covariance passed to the renderer are
\begin{equation}
    \begin{aligned}
        \mu_j^{\mathcal C}
        &=t_\star+R_\star x_j,\\
        \Sigma_j^{\mathcal C}
        &=R_\star F_j
        \operatorname{diag}\!\bigl((s_j^{\mathcal C})^2\bigr)
        F_j^\top R_\star^\top.
    \end{aligned}
    \label{eq:ew-supp-sidewall-world-parameters}
\end{equation}
We initialize sidewall opacity to $0.95$. Each base RGB color is initialized by bilinearly sampling the clean target image at the corresponding initial projected center, clipping the sample to $[0.02,0.98]$, and converting it to logits. The primitive count and orientation frames remain fixed, without densification or pruning. These bounds constrain Gaussian parameters without truncating their spatial tails. Numerical positive-definiteness safeguards are applied to the thin covariances.

\paragraph{SH evaluation.}
We use degree-three SH with 15 non-DC coefficients and optimize the attributes listed in Eq.~\eqref{eq:ew-main-cavity-parameters}. The compensated DC coefficient follows Eq.~\eqref{eq:ew-main-sh-compensation}, using the rendering convention in Eq.~\eqref{eq:ew-main-sh-color}. The target direction and its compensation are recomputed from each current center. The result can be encoded in ordinary SH coefficients without a target-camera-specific rendering branch. MVSplat and native optimization-based 3DGS use different SH basis conventions. Each adapter evaluates its own basis and gradients consistently. For native 3DGS, evaluated RGB values are supplied to the rasterizer as precomputed colors. Compensation controls each primitive's target-direction color before clipping, while changes in geometry, opacity, and compositing still require target-image supervision.

\subsection{Candidate Cameras and Fixed Supervision Views}
\label{sec:ew-supp-view-selection}

\subsubsection{Translation and orbit candidates}
To define the orbit center, retain original Gaussian centers within the
interval between the 2nd and 98th percentiles along each coordinate axis. Denoting their
index set by $\mathcal I$, the opacity-weighted center is
\begin{equation}
 c_{\rm orbit}=\frac{\sum_{i\in\mathcal I}\alpha_i\mu_i}
                         {\sum_{i\in\mathcal I}\alpha_i}.
 \label{eq:ew-supp-orbit-center}
\end{equation}
The local camera set contains the target, 36 translated cameras, and
60 orbit cameras, totaling 97. For translations, use pixel-equivalent
steps $s\in\{0.25,0.5,1,2,4,8\}$ and midpoint depth $z_m=0.33$:
\begin{equation}
 \delta=\pm\frac{sz_m}{f_x^\star},\qquad
 t'=t_\star+\delta R_\star e_a,\qquad a\in\{x,y,z\}.
 \label{eq:ew-supp-camera-translation}
\end{equation}
Here $e_a$ is a coordinate-axis unit vector and orientation stays fixed.
The implementation uses $f_x^\star$ for all three axes: a $z$ translation
is not an image shift of $s$ pixels, and the $y$ step is only exactly
pixel-equivalent when the two focal lengths match.

$\pm\{0.0025,0.005,0.01,0.02,0.05,0.1,0.25,0.5,1,2\}^{\circ}$ are the signed orbit angles. For yaw, pitch, and diagonal motion, respectively, form
\begin{equation}
 \begin{aligned}
 Q(\alpha)&\in\{R_y(\alpha),R_x(\alpha),R_y(\alpha)R_x(\alpha)\},\\
 A_{\rm rot}&=R_\star Q(\alpha)R_\star^\top,\\
 t'&=c_{\rm orbit}+A_{\rm rot}(t_\star-c_{\rm orbit}),
 \quad R'=A_{\rm rot}R_\star.
 \end{aligned}
 \label{eq:ew-supp-camera-orbit}
\end{equation}
Both position and orientation rotate around the scene center. For the
diagonal case, $\alpha$ is each component angle, not the total rotation
angle on $\mathrm{SO}(3)$.

\subsubsection{Perturbation scores and distance normalization}
Selection occurs after PCL and before sidewall color calibration.
Let $\widehat Y_k^{\rm sel}$ be the augmented rendering at this stage.
Candidates must have nonempty standalone added-structure support and a
maximum absolute RGB change above $\epsilon_{\rm RGB}=1/255$.
The perturbation score used in Eq.~\eqref{eq:ew-view-selection} is
\begin{equation}
 E_k=\sum_{p\in\Omega}\frac{1}{3}
       \|\widehat Y_k^{\rm sel}(p)-Y_k(p)\|_2^2.
 \label{eq:ew-supp-perturbation-score}
\end{equation}
It sums over pixels rather than averaging over the image.
For the diversity in Eq.~\eqref{eq:ew-view-diversity}, define
\begin{equation}
 \begin{aligned}
 d_t(k,\ell)&=\|t_k-t_\ell\|_2,\\
 d_R(k,\ell)&=\arccos\!\left(
 \operatorname{clip}\!\left(\frac{\operatorname{tr}(R_k^\top R_\ell)-1}{2},
                          -1,1\right)\right),\\
 \widehat d_a(k,\ell)&=
 \frac{d_a(k,\ell)}{\max(10^{-8},\max_{i,j\in\mathcal I_{\rm cand}}d_a(i,j))},
 \quad a\in\{t,R\}.
 \end{aligned}
 \label{eq:ew-supp-camera-distances}
\end{equation}
The weights in Eq.~\eqref{eq:ew-view-diversity} are
$(w_t,w_R,w_P)=(0.4,0.3,0.3)$. The masks in that equation are the initial standalone
watermark/sidewall support unions. Eq.~\eqref{eq:ew-view-selection}
uses $\lambda=0.5$, $K_{\rm train}=8$, and a $10^{-12}$ lower bound on
$E_{\max}$ in the denominator. The first camera maximizes $E_k$.
Subsequent selections never reuse a camera. Fewer than eight eligible
cameras are all retained. In Eq.~\eqref{eq:ew-view-selection}, $\mathcal S$ contains the non-target
views selected so far. Once selection is complete, $\mathcal S$ is held
fixed and used to compute the non-target loss in
Eq.~\eqref{eq:ew-main-nontarget-loss} at every BO step. With no eligible non-target camera, only target
optimization is performed and the empty selection is recorded.

\subsection{Color Learning and Sidewall Calibration}
\label{sec:ew-supp-initialization-optimization}

\subsubsection{PCL color initialization and color-space definitions}
\label{sec:ew-supp-watermark-calibration}
PCL uses the contrast and color-difference quantities in
Eq.~\eqref{eq:pco}, the objective in Eq.~\eqref{eq:pco-loss}, and the
watermark-only renderer in Eq.~\eqref{eq:pco-rendering}.
Each watermark Gaussian starts from the same clean-image mean RGB:
\begin{equation}
 \bar c_j^{\mathcal W,(0)}=\frac{1}{HW}\sum_{p\in\Omega}
             \operatorname{clip}(Y_\star(p),0,1).
 \label{eq:ew-supp-mean-color}
\end{equation}
Its RGB is subsequently optimized independently. The background set is
the entire $8\times8$ glyph box, including stroke locations and
low-coverage black pixels. Thus the IBM \texttt{A} with 28 active pixels
uses all $28\times64=1792$ foreground-background pairs. There is no pair
sampling or alpha-based background rejection.

Before evaluating the color objective, foreground and background RGB
values are clipped to $[0,1]$. For an encoded sRGB component $\mathbf{C}$, define
\begin{equation}
 g(c)=\begin{cases}
 c/12.92,&c\leq0.04045,\\
 ((c+0.055)/1.055)^{2.4},&c>0.04045.
 \end{cases}
 \label{eq:ew-supp-srgb-linearization}
\end{equation}
The relative luminance in Eq.~\eqref{eq:pco} is
\begin{equation}
 L_{\rm rel}(c)=0.2126g(c_R)+0.7152g(c_G)+0.0722g(c_B).
 \label{eq:ew-supp-relative-luminance}
\end{equation}
For the CIEDE2000 term, unit parametric factors
$k_L=k_C=k_H=1$ give
\begin{equation}
 \begin{aligned}
 \Delta E_{00}=\Bigg[
 &\left(\frac{\Delta L'}{S_L}\right)^2
 +\left(\frac{\Delta C'}{S_C}\right)^2
 +\left(\frac{\Delta H'}{S_H}\right)^2\\
 &+R_T\frac{\Delta C'}{S_C}\frac{\Delta H'}{S_H}
 \Bigg]^{1/2}.
 \end{aligned}
 \label{eq:ew-supp-ciede2000}
\end{equation}
Here $\Delta L'$, $\Delta C'$, and $\Delta H'$ are the corrected lightness,
chroma, and hue differences. $S_L,S_C,S_H$ are their weighting functions,
and $R_T$ is the chroma-hue interaction correction of CIEDE2000.

In Eqs.~\eqref{eq:pco} and \eqref{eq:pco-loss}, we set
$\epsilon_{\rm lum}=0.05$, $\tau_{\rm cr}=4.5$, and
$\lambda_{\rm cr}=\lambda_{\rm de}=1$.
Adam runs for 50 steps at learning rate of 0.02, clipping Gaussian RGB to
$[0,1]$ after each update. The lowest-loss state, including the initial
state, is retained before freezing the entire watermark. A 900-second
safety timeout is an error rather than acceptance of a partial PCL run.
These objectives promote color separation. Neither a contrast threshold
nor a color-loss minimum directly establishes correct letter recognition.

\subsubsection{Reference images and sidewall color calibration}
\label{sec:ew-supp-cavity-color-calibration}
The learned watermark-only target image is
$\widetilde Y_\star(\zeta^\star)$ from Eq.~\eqref{eq:pco-rendering}.
The target image $J_\star$ in Eq.~\eqref{eq:ew-main-target-image}
uses these colors on $M$ and the clean rendering elsewhere.
BO supervises the strokes against this same learned image and uses
$Y_\star$ for its background term, as in
Eq.~\eqref{eq:ew-main-nontarget-loss}.
Sidewall color calibration instead compares against the \emph{full}
watermark-only image:
\begin{equation}
 \mathcal L_{\rm cal}
 =\ell_{D_{\rm cal}}\!\left(
 \widehat Y_\star(\zeta^\star,\phi),
 \widetilde Y_\star(\zeta^\star)\right).
 \label{eq:ew-supp-cavity-calibration-loss}
\end{equation}
Here $D_{\rm cal}$ is the initial standalone sidewall projection and
$\ell_D$ is defined in Eq.~\eqref{eq:ew-main-masked-error}.
Only the base RGB colors of both sidewall layers are updated.
Geometry, opacities, and the learned watermark remain fixed.
The learning rate is 0.05, with at most 200 steps or 20 seconds.
Calibration may be skipped if the initial bright-edge diagnostic passes.
Saved states are ranked lexicographically by stroke-color coverage,
bright-edge pass status, and negative calibration MSE, rather than by
MSE alone. A timeout can continue to BO but is not a calibration pass.

The bright-edge proxy compares with the channelwise maximum of the clean
and watermark-only target images. In the aperture boundary ring, a pixel
is flagged if all three channels increase by more than $16/255$.
A horizontal or vertical flagged run of at least three pixels fails the
proxy. This diagnostic triggers calibration and ranks its checkpoints.
It is not an additional BO loss or a test of perceptual invisibility.
State $\mathbf B$ is saved after this calibration.

\subsection{BO Implementation and Numerical Safeguards}
\label{sec:ew-supp-main-optimization}
\label{sec:ew-supp-dynamic-regions}
\label{sec:ew-supp-numerical-adapters}

\subsubsection{Masks, objective, and checkpoints}
The region in Eq.~\eqref{eq:ew-main-dynamic-region} uses initial supports
recorded after sidewall color calibration, current supports, and every
pixel with an actual nonzero RGB difference from the clean rendering.
Standalone support is measured with white added Gaussians on black
background, retaining positive rasterized alpha. The difference test
uses no $1/255$ threshold. That threshold applies only to candidate
selection. The full stroke mask is additionally included for the target.
The region is recomputed but excluded from differentiation.

The objective and region normalization are given in
Eq.~\eqref{eq:ew-main-masked-error} through Eq.~\eqref{eq:ew-main-objective}.
We use $\lambda_{\rm tar}=50$, $\lambda_{\rm non}=500$, and
$\lambda_{\rm out}=1$, with Adam at a learning rate of 0.01 for 600 steps.
Each step accumulates gradients from the target and all selected
non-target cameras before one parameter update. BO uses floating-point
renderings directly. The clipping used by image evaluation is separate.
No LPIPS, SSIM, volume, or collision penalty is added to this objective.
A tenfold coefficient ratio does not imply a tenfold gradient contribution
because regions and residual magnitudes differ.

\subsubsection{Projection, covariance, and renderer checks}
Watermark centers are verified by the renderer's matrix projection and an
independent double-precision projection. Their warning thresholds are
0.002 and 0.001 pixels, respectively. Both have a hard limit of 0.2 pixels.
Invalid depth, non-finite values, or a hard-limit violation are errors.
These thresholds concern geometric anchors, not RGB preservation.
Products of rotations and scales, along with covariance Gram matrices, are constructed in
double precision with numerical positive-definiteness protection.
Parameters, gradients, and scales are checked for numerical failures.
Hashes of the original scene, frozen watermark, and reference image
verify that BO modifies only the sidewalls. SH basis order and coordinate conventions must match the renderer used
in Eqs.~\eqref{eq:ew-main-sh-color} and \eqref{eq:ew-main-sh-compensation}.
The MVSplat rasterizer and native Graphdeco 3DGS adapter use their own
basis implementations and corresponding derivatives. Their coefficient
indices cannot be interchanged without conversion.
For native 3DGS scenes, SH colors and their gradients are computed before
passing RGB to the rasterizer. These adapters preserve the common
optimization formulation without assuming pixel-identical outputs across
renderers.

\subsection{Evaluation Definitions}
\label{sec:ew-supp-evaluation}
\label{sec:ew-supp-image-evaluation}

\subsubsection{Evaluation states and common reference}
\label{sec:ew-supp-evaluation-states}
We compare states $\mathbf{A}$, $\mathbf{B}$, and $\mathbf{C}$ at identical cameras and image resolutions, using the clean rendering in Eq.~\eqref{eq:3dgs_encoder_decoder} as their common reference. State $\mathbf{A}$, denoted by $Y_k^A$, contains the original scene and the learned, frozen watermark without sidewalls. Its target rendering is Eq.~\eqref{eq:pco-rendering} evaluated at $\zeta^\star$. For the augmented rendering in Eq.~\eqref{eq:ew-main-augmented-rendering}, we write $Y_k^B=\widehat Y_k(\zeta^\star,\phi_0)$ and $Y_k^C=\widehat Y_k(\zeta^\star,\phi^\star)$. Here, $\phi_0$ is the sidewall state after target-direction color calibration and before BO, and $\phi^\star$ is the exported optimized state. Thus, $\mathbf{B}$ is the calibrated step-zero state, rather than the uncalibrated geometric template. The original scene and watermark are identical across all three states. Differences between their clean-reference scores quantify changes in fidelity. They are not direct image distances between states.

\subsubsection{Standalone projection supports and events}
\label{sec:ew-supp-projection-definitions}
For camera $P_k$, render the watermark and each sidewall state independently in white against black, without the original scene. Let $\alpha_{W,k}$ and $\alpha_{C,k}^X$ be the resulting accumulated alpha images. Their finite rasterized supports are
\begin{equation}
\begin{aligned}
 S_W(P_k)&=\{p:\alpha_{W,k}(p)>0\},\\
 S_C^X(P_k)&=\{p:\alpha_{C,k}^X(p)>0\},\quad X\in\{\mathbf{B},\mathbf{C}\}.
\end{aligned}
\label{eq:ew-supp-isolated-supports}
\end{equation}
The watermark support is shared by all states, and $S_C^A=\varnothing$. These supports include renderer clipping, truncation, filtering, and alpha thresholds. Although a mathematical Gaussian has infinite tails, zero rasterized support means that no pixel receives a contribution under these renderer rules. Standalone support does not account for occlusion by the original scene or establish glyph readability. We define the projection events as
\begin{equation}
\begin{aligned}
 P_W(P_k)&=\mathbf1[|S_W(P_k)|>0],\\
 P_{\mathrm{either}}^X(P_k)
   &=\mathbf1[|S_W(P_k)\cup S_C^X(P_k)|>0],\\
 P_{\mathrm{both}}^X(P_k)
   &=\mathbf1[|S_W(P_k)|>0\ \land\ |S_C^X(P_k)|>0],\\
 P_{\mathrm{cavityOnly}}^X(P_k)
   &=\mathbf1[|S_W(P_k)|=0\ \land\ |S_C^X(P_k)|>0].
\end{aligned}
\label{eq:ew-supp-projection-events}
\end{equation}
Here, $\mathbf1[\cdot]$ is the indicator function. ``Both'' means that each component has nonempty support. Their supported pixels need not intersect. Watermark-only and neither-present events follow by requiring, respectively, nonempty watermark and empty sidewall support, or both supports empty. The fraction of image pixels covered by either component is
\begin{equation}
 a_k^X=\frac{|S_W(P_k)\cup S_C^X(P_k)|}{H_kW_k}.
\label{eq:ew-supp-projected-area-fraction}
\end{equation}

\subsubsection{View groups and fixed comparison sets}
\label{sec:ew-supp-view-groups}
Target is the selected camera $P_\star$. Seen and Unseen are local non-target candidates included in and excluded from BO, respectively. These names do not refer to reconstruction inputs or unseen scenes. Local non-target evaluation considers all 96 candidate cameras,
partitioned into Seen and Unseen according to their use in
optimization. Local-crop metrics retain only cameras with a
valid projected watermark center and a complete crop,
using the same retained set across $\mathbf{A}/\mathbf{B}/\mathbf{C}$. The optimization-based 3DGS protocol additionally evaluates all $97$ local cameras and records initial projection flags. These sets retain their own denominators. A comparison conditioned on any affected state can use the fixed union
\begin{equation}
\begin{aligned}
 \mathcal P_{\mathrm{aff}}=\bigl\{P_k\in\mathcal P_{\mathrm{eval}}:\\
 |S_W(P_k)\cup S_C^B(P_k)\cup S_C^C(P_k)|>0\bigr\},
\end{aligned}
\label{eq:ew-supp-fixed-affected-set}
\end{equation}
where $\mathcal P_{\mathrm{eval}}$ is the specified evaluation camera set. The native-frame conditional fidelity in the main paper uses a different, explicitly stated condition:
\begin{equation}
 \mathcal P_{C,\mathrm{native}}
 =\{P_k\in\mathcal P_{\mathrm{native}}:
        P_{\mathrm{either}}^C(P_k)=1\}.
\label{eq:ew-supp-final-native-set}
\end{equation}
It selects cameras with remaining support at $\mathbf{C}$. Any accompanying $\mathbf{A}/\mathbf{B}/\mathbf{C}$ comparison uses this same selected set for every state. Neither condition permits independently selecting a favorable subset for each state. All-camera, initial-projection, union-affected, and final-$\mathbf{C}$ conditional results describe different populations and must retain their stated denominators. In a cavity-only view, the state-$\mathbf{A}$ rendering may be identical to the
clean rendering, giving zero MSE and infinite PSNR for that individual
image. 

\paragraph{PSNR aggregation.}
All aggregate PSNR values reported in the main paper and
supplementary material are computed from the arithmetic mean
of per-image mean squared errors (MSEs), rather than by
averaging individual PSNR values in decibels.
For a nonempty evaluation set of $N$ images or valid crops
with RGB values in $[0,1]$, we compute
\begin{equation}
\mathrm{PSNR}_{\mathrm{agg}}
=
-10\log_{10}
\left(
\frac{1}{N}\sum_{i=1}^{N}\mathrm{MSE}_i
\right),
\end{equation}
where $\mathrm{MSE}_i$ is averaged over all RGB samples
within the evaluated image or crop.
Each image or crop receives equal weight, and $N$ is
determined by the inclusion criteria of the corresponding
evaluation protocol.
Zero-error samples remain included; the aggregate PSNR
is infinite only when all evaluated samples have zero error.
SSIM and LPIPS, where reported, are averaged over individual
images or crops.
Within each $\mathbf{A}/\mathbf{B}/\mathbf{C}$ comparison, the same retained camera set
and, for local evaluation, the same crop regions are used.
\subsubsection{Target color preservation}
\label{sec:ew-supp-target-evaluation}
\label{sec:ew-supp-fidelity-and-projection}
\label{sec:ew-supp-target-checks}
Let $M$ be the target stroke mask. Its contour is $\partial M=M\setminus\operatorname{erode}_{3\times3}(M)$, where erosion uses a $3\times3$ structuring element. For each stroke pixel, define
\begin{equation}
 m(p)=\mathbf1\!\left[
   \|Y_\star^C(p)-Y_\star^A(p)\|_\infty\leq0.05
 \right],\qquad p\in M.
\label{eq:ew-supp-target-pixel-error}
\end{equation}
The infinity norm is the maximum absolute RGB-channel difference. The two coverage scores are
\begin{equation}
 q_{\mathrm{stroke}}=\frac{\sum_{p\in M}m(p)}{|M|},
 \qquad
 q_{\mathrm{contour}}=\frac{\sum_{p\in\partial M}m(p)}{|\partial M|}.
\label{eq:ew-supp-target-coverage}
\end{equation}
Both scores must be at least $0.90$. Thus, the current $28$-pixel letter \texttt{A} requires at least $26$ qualifying stroke pixels and must also pass the contour test. Other glyphs use their own stroke and contour cardinalities. On $M$, the reference $Y_\star^A$ equals the learned target reference in Eq.~\eqref{eq:ew-main-target-image}. 

\subsubsection{Rotational projection scans}
\label{sec:ew-supp-projection-scans}
Projection scans keep target intrinsics fixed and orbit the camera around the Gaussian scene center in positive and negative yaw, pitch, and diagonal directions. Their sampling lists are independent of the rotational candidates used to choose BO supervision views. A diagonal defined by equal yaw and pitch component angles differs from one labeled by its total rotation angle. For a desired signed total angle $\theta$, an equal-component construction is
\begin{equation}
\begin{aligned}
 \varphi&=2\arcsin\!\left(\sqrt2\sin\frac{|\theta|}{4}\right),\\
 Q(\theta)&=R_y(\operatorname{sgn}(\theta)\varphi)
           R_x(\operatorname{sgn}(\theta)\varphi),
\end{aligned}
\label{eq:ew-supp-diagonal-total-angle}
\end{equation}
where $R_x$ and $R_y$ are rotations about the local camera axes and trigonometric arguments are in radians. The scan records whether its angle denotes a component angle or the total rotation angle. For the optimized scene, the projection pixel counts are
\begin{equation}
 n_W(P_k)=|S_W(P_k)|,\qquad n_C(P_k)=|S_C^C(P_k)|.
\label{eq:ew-supp-projection-counts}
\end{equation}
Watermark, cavity, and joint disappearance correspond to $n_W=0$, $n_C=0$, and $n_W=n_C=0$, respectively. Along a fixed scan direction, if consecutive sampled offset magnitudes $\beta_{q-1}<\beta_q$ change from nonempty to empty support, the observed transition is bracketed by
\begin{equation}
 \beta_{\mathrm{vanish}}\in(\beta_{q-1},\beta_q].
\label{eq:ew-supp-disappearance-interval}
\end{equation}
We distinguish the first disappearance interval, the interval after which all subsequent sampled projections remain empty, and any reappearance. No monotonicity is assumed. If disappearance is not observed by the scan limit, it remains unobserved within that range. It is not assigned the limit angle. Empty projection can result from leaving the frame, clipping, or renderer thresholds and alone does not demonstrate cavity occlusion.

\subsubsection{Original-camera sampling and aggregation}
\label{sec:ew-supp-original-camera-evaluation}
Dataset cameras retain their original poses. Native frames use their original intrinsics and resolutions, whereas $256$-pixel crops use the corresponding crop-adjusted intrinsics. Neither applies the target focal enlargement. In the MVSplat random-$32$ protocol, the sampling pool for scene $s$ is
\begin{equation}
\begin{aligned}
 \mathcal P_{s,\mathrm{pool}}
   &=\mathcal P_s^{\mathrm{data}}
      \setminus(\mathcal P_{s,\mathrm{ref}}\cup\mathcal P_{s,\mathrm{start}}),\\
 n_s&=\min(32,|\mathcal P_{s,\mathrm{pool}}|).
\end{aligned}
\label{eq:ew-supp-original-camera-pool}
\end{equation}
Here, $\mathcal P_{s,\mathrm{ref}}$ and $\mathcal P_{s,\mathrm{start}}$ contain the original dataset records selected as placement references and the search starting camera. Exclusion is performed by source-frame identity before changing intrinsics, rather than by comparing native and crop-adjusted camera matrices. A saved random seed selects $n_s$ cameras without replacement. Native and crop evaluations use the same frames and extrinsics. Excluding these cameras does not necessarily exclude all reconstruction inputs.

Full-camera evaluation is a separate protocol. MVSplat full-camera manifests include reference, starting, and reconstruction-input frames, while also marking the subset that excludes placement references and starting frames. The optimization-based 3DGS evaluation uses all registered COLMAP cameras for MIP and the deduplicated train/test union for NS. Its random-$32$ flags follow its own camera-table sampling rather than automatically inheriting the MVSplat exclusion rule. Each result uses its actual evaluated list, including when fewer than $32$ eligible cameras exist. No camera is automatically discarded because the original rendering has low visual quality.

For the declared evaluated set $\mathcal P_s^{\mathrm{eval}}$ of each scene, final projection incidence is
\begin{equation}
 r_{\mathrm{projection}}^C
 =\frac{\displaystyle\sum_s\sum_{P_k\in\mathcal P_s^{\mathrm{eval}}}
             P_{\mathrm{either}}^C(P_k)}
        {\displaystyle\sum_s|\mathcal P_s^{\mathrm{eval}}|}.
\label{eq:ew-supp-projection-occurrence-rate}
\end{equation}
The denominator is the actual camera count, not the number of scenes times a nominal sampling budget. We separately record both-present, watermark-only, cavity-only, and neither-present events. These are standalone projection statistics, not human visibility rates in composited images.

\subsubsection{Conditional hiding and interpretation}
\label{sec:ew-supp-cavity-effectiveness}
Target color preservation, image fidelity, projection incidence, and recognition answer different questions. Comparing $\mathbf{B}$ and $\mathbf{C}$ measures the change from optimizing an existing cavity. Comparing $\mathbf{A}$ and $\mathbf{C}$ measures the net fidelity effect of adding and optimizing it. Neither full-image fidelity nor projection absence alone establishes unreadability.

A conditional hiding measure requires a visibility or readability rule fixed before evaluation. Let $V_{\rm eval}(Y,Y_k)\in\{0,1\}$ indicate whether a letter remains visible or readable under that rule, using $Y_k$ as the clean control. Let $\mathcal P_{A\text{-visible}}$ contain matched cameras where the watermark-only rendering differs from the clean image and passes the declared baseline visibility criterion. Then
\begin{equation}
 r_{\mathrm{hide}}
 =\frac{\displaystyle\sum_{P_k\in\mathcal P_{A\text{-visible}}}
             \mathbf1[V_{\rm eval}(Y_k^C,Y_k)=0]}
        {|\mathcal P_{A\text{-visible}}|}.
\label{eq:ew-supp-proposed-hiding-rate}
\end{equation}
This operational definition complements the ideal condition in
Eq.~\eqref{eq:watermark_formulation}. Projection counts and fidelity scores
do not supply the required recognition rule. The before/after comparison
uses the same cameras. Under a strict hiding protocol, only an absent or
uncertain letter is counted as suppressed. A wrong-letter transcription
is recorded separately rather than automatically treated as hiding.
Consequently, hiding is not simply one minus correct-glyph accuracy.
An empty conditioning set has no evaluable samples and is not assigned
perfect hiding. Statistical uncertainty across multiple views of a scene
should account for their within-scene dependence.
\subsubsection{MLLM-based recognition evaluation}
\label{sec:supp-recognition-protocol}

Unless otherwise specified, recognition is evaluated using
\texttt{gpt-6-luna} with high reasoning effort and high image
detail. Each request provides one evaluated rendering and
an IBM font reference, without the clean rendering or the
ground-truth letter. We use the same prompt, which directs the evaluator to focus on inspecting the four image corners without revealing the watermark's exact location.
Responses are categorized as the intended letter, another
letter, uncertain, or none. Recognition accuracy is the
number of correct intended-letter predictions divided by
the number of evaluated images; wrong-letter, uncertain,
and none responses are all counted as incorrect.
The intended letter is used only for scoring after inference.
Within each matched $\mathbf{A}/\mathbf{B}/\mathbf{C}$ comparison, the same camera
positions are evaluated at every stage.

Conditional suppression is evaluated separately on images
correctly recognized at $\mathbf{A}$ and measures the fraction no longer
correctly recognized at $\mathbf{C}$. It includes wrong-letter responses.
Strict hiding instead counts only uncertain or none responses
within the same A-correct subset.
\section{Post-hoc Hypothesis Tests}
\label{sec:hypothesis-tests}

We report a model-restricted post-hoc analysis of MVSplat and
IDEsplat on RE10K and ACID, with a separate 26-glyph
MVSplat/RE10K experiment. Due to the limited number of scenes in the MIP and NS datasets, the optimization-based 3DGS cohorts are excluded from these statistical analyses.
All experiments use adaptive watermark colors and
target-camera focal lengths equal to $2.5$ times the
per-scene maxima. Recognition uses the same prompt, with an IBM font reference and no clean reference image.


\paragraph{Scope of ownership verification.} Ownership verification refers to inspecting a marked\
asset using the retained target camera $P_\star$ and the expected\
glyph. The threshold and offsets were selected after inspecting the recognition results. Reading the expected glyph at that camera provides supporting\
evidence for an attribution claim. The analyses below evaluate two\
properties relevant to this inspection: target-view readability and\
suppression of correct transcription at sampled neighboring views.\
\
\paragraph{Post-hoc analysis.}\
For the four feed-forward model--dataset settings, we analyze the
same scene-level recognition records used in Table~\ref{tab:exp-recognition}. These post-hoc\
tests quantify the evidence for target-view readability and for a\
reduction in correct transcription after adding the optimized cavity\
at matched non-target views. We also report a separate post-hoc\
threshold analysis for the 26-glyph experiment at the sampled\
$1^\circ$ and $2^\circ$ offsets. The reported $p$-values concern\
these recognition outcomes. The probability of accepting a competing\
ownership claim remains unmeasured.
\subsection{Scene and View Sampling}
\label{sec:hypothesis-sampling}

\paragraph{Four model--dataset settings.}
The feed-forward recognition results in the main paper
(Table~\ref{tab:exp-recognition}) and the hypothesis tests below
use the same sampled scenes and recognition records.
Using a fixed random seed, we sample approximately 10\%
of the completed scenes from each setting:
647 of 6,474 for MVSplat/RE10K,
160 of 1,595 for MVSplat/ACID,
100 of 1,000 for IDEsplat/RE10K, and
160 of 1,595 for IDEsplat/ACID.
Scene selection does not depend on recognition outcomes.
For each scene, we evaluate its selected target camera,
one randomly sampled non-target camera used during cavity
optimization (Seen), and six randomly sampled cameras
excluded from optimization (Unseen).
The Unseen pool comprises 18 rotations about the scene
Gaussian center: yaw, pitch, and diagonal rotations in both
directions at $0.05^\circ$, $0.1^\circ$, and $0.25^\circ$.
We exclude all training positions before sampling six
cameras from the remaining pool.
These are synthetic nearby rotations, rather than sampled
original dataset cameras, and contain no translation.
Stages $\mathbf{A}$ and $\mathbf{C}$ are evaluated at identical camera poses.

\paragraph{Separate 26-glyph experiment.}
We randomly select 50 shared scenes from those with completed
training for all 26 glyphs, without filtering by recognition
outcomes. Every glyph uses the same scene set.
The original scan evaluates yaw, pitch, and diagonal rotations
in both directions at $0.05^\circ$, $0.1^\circ$, $0.25^\circ$,
$0.5^\circ$, $1^\circ$, and $2^\circ$.
The threshold analysis outside the permitted observation
region uses only the tested $1^\circ$ and $2^\circ$ magnitudes,
giving $50\times26\times2\times3\times2=15{,}600$
scene--glyph--camera settings.
Conditional suppression is evaluated on settings where
stage $\mathbf{A}$ is correctly recognized, using the matched stage-$\mathbf{C}$
image. We additionally report an Unseen-only analysis,
excluding positions used to optimize the corresponding
scene--glyph model.
\subsection{Target-view readability}
For glyph A, one-sided exact binomial tests assess
$H_0:p_{\mathrm{target}}\leq0.90$, with one stage-$\mathbf{C}$ target
image per scene and Holm correction across four settings.
Table~\ref{tab:hyp-target-ff} supports recognition above 90\%
for MVSplat/RE10K, MVSplat/ACID, and IDEsplat/ACID
(adjusted $p<0.01$).
IDEsplat/RE10K achieves 91.00\%, but the test does not
establish a population recognition rate above 90\%
($p=0.4513$).

\begin{table*}[t]
\centering
\caption{Target-view recognition of glyph A at stage $\mathbf{C}$.
One-sided exact binomial tests assess
$H_0:p_{\mathrm{target}}\leq0.90$.
Holm correction covers the four model--dataset settings.
Intervals are marginal two-sided 95\% Clopper--Pearson
confidence intervals.}
\label{tab:hyp-target-ff}
\small
\begin{tabular}{lrrrr}
\toprule
Model / Dataset & Correct / Scenes & Acc. (\%) &
95\% CI (\%) & Holm-adjusted $p$ \\
\midrule
MVSplat / RE10K  & 617/647 & 95.36 & 93.45--96.85
& $1.65\times10^{-6}$ \\
MVSplat / ACID   & 158/160 & 98.75 & 95.56--99.85
& $2.52\times10^{-5}$ \\
IDEsplat / RE10K & 91/100  & 91.00 & 83.60--95.80
& 0.4513 \\
IDEsplat / ACID  & 154/160 & 96.25 & 92.02--98.61
& 0.0056 \\
\bottomrule
\end{tabular}
\end{table*}

\subsection{Paired recognition suppression}
At matched camera poses, we compute each scene's difference
between its $\mathbf{A}$- and $\mathbf{C}$-stage recognition rates and use the
unweighted mean of these differences as the test statistic.
Exact scene-block label permutations swap A and C jointly
across all views within each scene.
This test assumes independent scene clusters and
stage-label exchangeability under the null.
Holm correction covers eight Seen/Unseen comparisons.
All eight show reduced correct recognition
(adjusted $p<0.001$; Table~\ref{tab:hyp-paired-ff}).
Conditional suppression is the fraction of A-correct images
that are no longer correctly recognized at C, including
wrong-letter predictions. It ranges from 98.11\% to 99.18\%
on Unseen nearby rotations.
The views in this comparison use angular offsets of
$0.05^\circ$, $0.1^\circ$, and $0.25^\circ$ from the target view.

\begin{table*}[t]
\centering
\caption{Matched A/C recognition comparisons for glyph A.
The $p$-values test a directional reduction in recognition
using exact scene-block label permutations, rather than
suppression above 90\%. Holm correction covers all eight
Seen/Unseen comparisons. Conditional suppression counts
A-correct images that are not correctly recognized at C,
including wrong-letter responses.
Intervals use 50,000 scene-cluster bootstrap resamples.
Unseen rotations here are $0.05^\circ$, $0.1^\circ$,
and $0.25^\circ$.}
\label{tab:hyp-paired-ff}
\small
\setlength{\tabcolsep}{4pt}
\begin{tabular}{llrrrr}
\toprule
Model / Dataset & Views &
Acc. $\mathbf{A}$ $\to$ $\mathbf{C}$ (\%) &
Suppressed / A-correct &
Supp. (\%) [95\% CI] &
Holm-adjusted $p$ \\
\midrule
MVSplat / RE10K & Seen
& 74.65$\to$1.24 & 475/483
& 98.34 [97.11--99.38] & $7.18\times10^{-143}$ \\
MVSplat / RE10K & Unseen
& 53.66$\to$0.75 & 2054/2083
& 98.61 [98.06--99.10] & $2.41\times10^{-181}$ \\
MVSplat / ACID & Seen
& 76.88$\to$3.75 & 117/123
& 95.12 [90.98--98.41] & $3.61\times10^{-35}$ \\
MVSplat / ACID & Unseen
& 30.10$\to$0.31 & 286/289
& 98.96 [97.69--100.00] & $9.24\times10^{-33}$ \\
IDEsplat / RE10K & Seen
& 76.00$\to$1.00 & 75/76
& 98.68 [95.77--100.00] & $2.65\times10^{-23}$ \\
IDEsplat / RE10K & Unseen
& 35.33$\to$0.67 & 208/212
& 98.11 [96.08--99.56] & $1.03\times10^{-25}$ \\
IDEsplat / ACID & Seen
& 72.50$\to$2.50 & 112/116
& 96.55 [92.92--99.18] & $7.70\times10^{-34}$ \\
IDEsplat / ACID & Unseen
& 25.52$\to$0.21 & 243/245
& 99.18 [97.92--100.00] & $6.02\times10^{-35}$ \\
\bottomrule
\end{tabular}
\end{table*}

\subsection{Separate 26-glyph threshold analysis}
We separately analyze all 26 glyphs on the same 50 scenes
at the tested $1^\circ$ and $2^\circ$ rotation magnitudes.
For scene $i$, let $N_i$ denote the number of A-correct
positions and $S_i$ the number no longer correctly recognized
at C. To test $H_0:q\leq0.90$, where
$q=\mathbb{E}[S_i]/\mathbb{E}[N_i]$, we apply a one-sided
scene-level $t$ approximation to $U_i=S_i-0.90N_i$,
retaining scenes with $N_i=0$.
Holm correction covers the four contrasts in
Table~\ref{tab:hyp-orbit26}.
Pooled conditional suppression is 99.90\%
(2,879/2,882; 95\% CI: 99.73--100.00\%).
Excluding training positions yields 99.88\%
(844/845; 95\% CI: 99.58--100.00\%).
All four approximate threshold tests have adjusted
$p<0.001$. The Unseen-only subset is skewed, so its
$t$ approximation and extreme-tail probabilities require
cautious interpretation. These findings concern the
measured magnitudes and do not establish invisibility
throughout an unmeasured continuous range.

\begin{table*}[t]
\centering
\caption{Separate 26-glyph experiment on 50 shared
MVSplat/RE10K scenes, using yaw, pitch, and diagonal rotations
in both directions. A one-sided scene-level $t$ approximation
tests $H_0:q\leq0.90$ using $U_i=S_i-0.90N_i$,
with 49 degrees of freedom.
Holm correction covers these four contrasts.
Intervals use 100,000 scene-cluster bootstrap resamples.
The pooled results include training positions unless marked
Unseen. Results concern the tested magnitudes, not an entire
continuous angular interval.}
\label{tab:hyp-orbit26}
\small
\begin{tabular}{lrrrr}
\toprule
Rotation subset &
Suppressed / A-correct &
Supp. (\%) &
95\% CI (\%) &
Holm $p$ (approx.) \\
\midrule
$1^\circ$ and $2^\circ$, pooled
& 2879/2882 & 99.90 & 99.73--100.00
& $4.74\times10^{-16}$ \\
$1^\circ$
& 1891/1892 & 99.95 & 99.83--100.00
& $3.02\times10^{-19}$ \\
$2^\circ$
& 988/990 & 99.80 & 99.36--100.00
& $8.95\times10^{-8}$ \\
$1^\circ$ and $2^\circ$, Unseen only
& 844/845 & 99.88 & 99.58--100.00
& $2.78\times10^{-6}$ \\
\bottomrule
\end{tabular}
\end{table*}

\subsection{Uncertainty and multiplicity}
Conditional suppression intervals use scene-cluster
percentile bootstrap resampling, keeping all glyphs, views,
and stages of each scene together.
Reported 95\% intervals are marginal, not simultaneous.
Holm adjustment is performed separately within the three
stated families of 4, 8, and 4 tests at $\alpha=0.05$.
This controls multiplicity within each family, rather than
providing a joint family-wise guarantee across all analyses.

\section{Additional Experiments}
\label{sec:additional-experiments}

We examine additional design choices concerning placement,
view selection, cavity geometry and parameter optimization,
glyph resolution, and rendering, and report complementary
visibility diagnostics.
Cavity layer count and PCL color loss weights are already
evaluated in Tables~\ref{tab:ablation-cavity-layers}
and~\ref{tab:color-loss} of the main paper and are not
repeated here.

\paragraph{Protocol and populations.}
Unless otherwise specified, the following ablations use the
same 100 MVSplat/RE10K scenes and IBM glyph A, including the
additional cavity-parameter, planar-patch, and glyph-resolution
studies; focal-length sensitivity uses separately specified
scene cohorts. This ablation cohort is distinct from the larger reconstruction and alphabet evaluations. Target recognition uses \texttt{6-luna} with high reasoning effort, the same prompt, and the IBM font blind-test protocol. Accuracy is measured over all 100 target images. Paired retention counts correctly recognized images within the fixed 96 scenes recognized by the baseline. The baseline recognition records are reused across sweeps, so its retention of 96/96 is a self-reference rather than an independent validation of the default.

Full-image non-target fidelity evaluates all 9,600 candidate
views, including optimization views and untrained views.
The additional $8\times8$ local metrics include only
positions admitting a complete valid crop, with identical
crops and eligible positions across $\mathbf{A}/\mathbf{B}/\mathbf{C}$ within each
configuration. The crop remains $8\times8$ for enlarged glyphs.
Projection incidence is evaluated separately on 15,261
original dataset cameras.

The cavity-length rotation experiment uses 20 scenes,
six magnitudes from $0.05^\circ$ to $2^\circ$, three
rotation modes, and both signs, giving 720 positions.
Its strict suppression uses the fixed 282 A-readable
images and counts only ``none'' or ``uncertain'' responses
at C, excluding wrong-letter predictions.

The additional planar-patch and glyph-resolution scans
use all 100 scenes, with 100 target positions and 600
positions at each of $0.25^\circ$ and $0.5^\circ$.
Their conditional suppression counts all incorrect
$\mathbf{C}$-stage responses among matched A-correct images,
including wrong letters.
At $0.25^\circ$ and $0.5^\circ$, we evaluate all sampled rotation
positions, including any positions used for BO supervision.
These positions are not restricted to Unseen positions.

State $\mathbf{A}$ is the original scene with the color-optimized watermark, state $\mathbf{B}$ adds the initial cavity, and state $\mathbf{C}$ follows cavity optimization. PSNR, SSIM, and LPIPS compare with a clean rendering at the same camera and resolution. These image metrics describe scene perturbation rather than watermark recognition. Unless specified otherwise, tables report state $\mathbf{C}$, recognition is evaluated after cavity optimization, and shaded rows identify defaults. The results are descriptive aggregates, without repeated-run uncertainty estimates. Shared and configuration-specific $A/B$ measurements are collected in Table~\ref{tab:supp-ablation-reference-states} to complement the final-state comparisons.

\subsection{Camera Retreat}
\label{sec:supp-ablation-retreat}

We compare placement search with camera retreat against insertion at the initial camera without retreat. Table~\ref{tab:supp-ablation-retreat} shows that retreat reduces final projection incidence from 56.798\% to 0.688\% under native framing and from 51.786\% to 0.557\% under the official crop. On the entire cropped camera collection, final PSNR increases from 38.04 to 53.97~dB and LPIPS decreases from 0.005166 to 0.000092. These measurements support retreat as a way to reduce exposure under the original dataset cameras. They do not measure how often a letter is recognizable.

The placement change already alters state $\mathbf{A}$, as shown by the complete stage measurements in the same table. This comparison therefore evaluates the placement procedure as a whole rather than isolating BO on an otherwise identical insertion. PSNR decreases from $\mathbf{B}$ to $\mathbf{C}$ in both conditions, while SSIM improves.

\begin{table*}[!tp]
\centering
\caption{Camera retreat on 15,261 original dataset cameras. A projection is nonempty standalone support from either the watermark or the cavity in state $\mathbf{C}$. Native and cropped incidence are separate measurements. Fidelity uses every camera with the official $256\times256$ crop, including cameras without projected support. The second panel lists states $\mathbf{A}/\mathbf{B}/\mathbf{C}$.}
\label{tab:supp-ablation-retreat}
{\fontsize{8.7}{10.5}\selectfont
\setlength{\tabcolsep}{5pt}
\renewcommand{\arraystretch}{1.12}
\begin{tabular}{@{}lrr@{}}
\toprule
Placement & Native count (\%) & $256\times256$ crop count (\%) \\
\midrule
Initial camera, no retreat & 8,668 (56.798) & 7,903 (51.786) \\
\rowcolor{gray!8} Search with retreat & 105 (0.688) & 85 (0.557) \\
\bottomrule
\end{tabular}

\medskip
\begin{tabular}{@{}lccc@{}}
\toprule
Placement & PSNR $\uparrow$ ($\mathbf{A}/\mathbf{B}/\mathbf{C}$) & SSIM $\uparrow$ ($\mathbf{A}/\mathbf{B}/\mathbf{C}$) & LPIPS $\downarrow$ ($\mathbf{A}/\mathbf{B}/\mathbf{C}$) \\
\midrule
Initial camera, no retreat & 46.56 / 38.20 / 38.04 & 0.999505 / 0.997593 / 0.997613 & 0.002449 / 0.004814 / 0.005166 \\
\rowcolor{gray!8} Search with retreat & 66.40 / 56.66 / 53.97 & 0.999995 / 0.999976 / 0.999978 & 0.000060 / 0.000088 / 0.000092 \\
\bottomrule
\end{tabular}}
\end{table*}

\subsection{Frustum Exclusion Criterion}
\label{sec:supp-ablation-qout}

The placement search uses the $q_{\mathrm{out}}$ test in Eq.~\eqref{eq:ew-main-placement-criteria} to favor locations outside the reference camera frusta. Removing this test increases final projection incidence in both framing conventions (Table~\ref{tab:supp-ablation-qout}). The count rises from 105 to 143 cameras under native framing and from 85 to 115 with the official crop. Thus, the frustum test further reduces geometric exposure within the same placement procedure. Nonempty support does not by itself imply a visible artifact or a recognizable letter.

\begin{table}[!htbp]
\centering
\caption{Effect of the $q_{\mathrm{out}}$ test on final projection incidence. Each row evaluates the same 15,261 dataset cameras. A camera is counted when either the watermark or the cavity has nonempty projected support.}
\label{tab:supp-ablation-qout}
{\fontsize{8.5}{10.2}\selectfont
\setlength{\tabcolsep}{3.0pt}
\renewcommand{\arraystretch}{1.12}
\begin{tabular}{@{}llrr@{}}
\toprule
Framing & Test & Cameras & Rate (\%) \\
\midrule
Native & Off & 143 & 0.937 \\
\rowcolor{gray!8} Native & On & 105 & 0.688 \\
$256\times256$ & Off & 115 & 0.754 \\
\rowcolor{gray!8} $256\times256$ & On & 85 & 0.557 \\
\bottomrule
\end{tabular}}
\end{table}

\subsection{Training View Selection}
\label{sec:supp-ablation-selection}

We compare three rules for selecting eight non-target supervision cameras. Combining leakage and view diversity gives the best final non-target fidelity among these rules, with PSNR of 37.25~dB compared with 37.07~dB for leakage ranking alone and 33.25~dB for random selection (Table~\ref{tab:supp-ablation-selection}). Target recognition is 96\%, 94\%, and 93\%, respectively. The gains over leakage ranking alone are modest.

Random selection has the highest target PSNR but lower recognition accuracy. A watermarked image closer to its clean reference need not contain a more readable letter. The separate color-preservation pass counts are 37, 34, and 0 out of 100. This diagnostic must not be read as a recognition score.

\begin{table*}[!tp]
\centering
\caption{View selection with eight non-target optimization views. State-$\mathbf{C}$ fidelity uses 9,600 candidates or 100 targets as indicated. Accuracy uses 100 targets and retention uses the fixed 96 baseline-recognized targets. Color counts passes of the separate color-preservation check. The three rules share states $\mathbf{A}$ and $\mathbf{B}$.}
\label{tab:supp-ablation-selection}
{\fontsize{8.1}{10.2}\selectfont
\setlength{\tabcolsep}{2.7pt}
\renewcommand{\arraystretch}{1.12}
\begin{tabular}{@{}lrrrrrrrrr@{}}
\toprule
 & \multicolumn{3}{c}{All non-target candidates} & \multicolumn{3}{c}{Target view} & \multicolumn{3}{c}{Target diagnostics} \\
\cmidrule(lr){2-4}\cmidrule(lr){5-7}\cmidrule(lr){8-10}
Selection & PSNR $\uparrow$ & SSIM $\uparrow$ & LPIPS $\downarrow$ & PSNR $\uparrow$ & SSIM $\uparrow$ & LPIPS $\downarrow$ & Acc. (\%) & Retained & Color \\
\midrule
\rowcolor{gray!8} Leakage + diversity & 37.25 & 0.996874 & 0.006982 & 36.42 & 0.997480 & 0.007311 & 96 & 96/96 & 37/100 \\
Largest leakage & 37.07 & 0.996846 & 0.007018 & 36.41 & 0.997461 & 0.007255 & 94 & 92/96 & 34/100 \\
Random & 33.25 & 0.995722 & 0.007727 & 38.94 & 0.997982 & 0.006219 & 93 & 92/96 & 0/100 \\
\bottomrule
\end{tabular}}
\end{table*}

\subsection{Number of Optimization Views}
\label{sec:supp-ablation-view-count}

We vary the number $n$ of non-target optimization views from 2 to 12 using leakage and diversity for selection. The target-preservation view remains active and is not included in $n$. Evaluation always uses the same 9,600 candidates. Final PSNR, SSIM, and LPIPS improve across the tested counts (Table~\ref{tab:supp-ablation-view-count}). PSNR gains 2.45~dB from two to eight views and a further 0.71~dB from eight to twelve, indicating smaller gains over the latter interval.

The color-preservation pass count decreases from 88/100 to 5/100. However, target recognition varies between 93\% and 97\% and does not decrease monotonically with $n$. The default of eight views is therefore a configuration choice rather than the optimum of every reported measure.

\begin{table*}[!tp]
\centering
\caption{Number of non-target optimization views. State-$\mathbf{C}$ fidelity uses the fixed 9,600 candidate views and 100 targets. Target accuracy and retention have denominators 100 and 96. Color is a separate pass count over 100 scenes. States $\mathbf{A}$ and $\mathbf{B}$ are shared across settings.}
\label{tab:supp-ablation-view-count}
{\fontsize{8.3}{10.2}\selectfont
\setlength{\tabcolsep}{3.0pt}
\renewcommand{\arraystretch}{1.12}
\begin{tabular}{@{}rrrrrrrrrr@{}}
\toprule
 & \multicolumn{3}{c}{All non-target candidates} & \multicolumn{3}{c}{Target view} & \multicolumn{3}{c}{Target diagnostics} \\
\cmidrule(lr){2-4}\cmidrule(lr){5-7}\cmidrule(lr){8-10}
$n$ & PSNR $\uparrow$ & SSIM $\uparrow$ & LPIPS $\downarrow$ & PSNR $\uparrow$ & SSIM $\uparrow$ & LPIPS $\downarrow$ & Acc. (\%) & Retained & Color \\
\midrule
2 & 34.80 & 0.996012 & 0.007729 & 36.49 & 0.997732 & 0.007004 & 97 & 95/96 & 88/100 \\
4 & 35.96 & 0.996443 & 0.007428 & 36.42 & 0.997638 & 0.007181 & 97 & 95/96 & 77/100 \\
6 & 36.66 & 0.996690 & 0.007199 & 36.40 & 0.997553 & 0.007257 & 93 & 91/96 & 60/100 \\
\rowcolor{gray!8} 8 & 37.25 & 0.996874 & 0.006982 & 36.42 & 0.997480 & 0.007311 & 96 & 96/96 & 37/100 \\
10 & 37.71 & 0.996995 & 0.006828 & 36.38 & 0.997429 & 0.007331 & 95 & 93/96 & 24/100 \\
12 & 37.96 & 0.997046 & 0.006722 & 36.40 & 0.997374 & 0.007298 & 96 & 94/96 & 5/100 \\
\bottomrule
\end{tabular}}
\end{table*}

\subsection{Cavity Length}
\label{sec:supp-ablation-length}

We hold the rear depth at 0.45 and set the front depth to 0.33, 0.30, 0.27, 0.24, or 0.21, giving cavity lengths from $0.5$ to $1.0$ times the default. Shorter cavities slightly reduce final non-target perceptual perturbation. For example, LPIPS is 0.006570 at $0.5\times$ and 0.006982 at $1\times$. This comes with weaker rotational suppression (Table~\ref{tab:supp-ablation-length}). Strict suppression increases from 194/282 (68.79\%) to 276/282 (97.87\%) over these settings. Target recognition ranges from 94\% to 98\%. In the rotation sweep, the reported wrong-letter counts are 19, 14, 14, 7, and 3 for increasing lengths.

The rotation sweep covers six sampled offsets from $0.05^\circ$ to
$2^\circ$. Correct-letter predictions at stage $\mathbf{C}$ indicate residual
readability at the corresponding camera positions. The correct-letter
count uses all 720 rotated views. Strict suppression uses the fixed
subset of 282 views with a correct stage-$\mathbf{A}$ prediction and counts
only ``none'' or ``uncertain'' responses at stage $\mathbf{C}$. The longer
default cavity favors suppression over the small fidelity advantage
of shorter configurations.

\begin{table*}[!tp]
\centering
\caption{Cavity length and recognition. Fidelity uses state $\mathbf{C}$ over 9,600 non-target candidates. Target accuracy uses 100 images and retention uses 96. Correct counts intended-letter predictions over all 720 rotation positions. Suppressed counts only none or uncertain responses within the 282 images readable in state $\mathbf{A}$, with the percentage in parentheses. Wrong-letter outputs are excluded from suppression.}
\label{tab:supp-ablation-length}
{\fontsize{8.5}{10.2}\selectfont
\setlength{\tabcolsep}{4.0pt}
\renewcommand{\arraystretch}{1.12}
\begin{tabular}{@{}rrrrrrrr@{}}
\toprule
 & \multicolumn{3}{c}{Non-target fidelity} & \multicolumn{2}{c}{Target recognition} & \multicolumn{2}{c}{Rotation sweep} \\
\cmidrule(lr){2-4}\cmidrule(lr){5-6}\cmidrule(lr){7-8}
Length & PSNR $\uparrow$ & SSIM $\uparrow$ & LPIPS $\downarrow$ & Acc. (\%) & Retained & Correct / 720 & Suppressed / 282 \\
\midrule
$0.5\times$ & 36.95 & 0.997341 & 0.006570 & 94 & 92/96 & 69 & 194 (68.79\%) \\
$0.625\times$ & 37.15 & 0.997285 & 0.006585 & 98 & 96/96 & 34 & 234 (82.98\%) \\
$0.75\times$ & 37.26 & 0.997194 & 0.006654 & 95 & 93/96 & 13 & 256 (90.78\%) \\
$0.875\times$ & 37.28 & 0.997049 & 0.006776 & 95 & 94/96 & 7 & 268 (95.04\%) \\
\rowcolor{gray!8} $1\times$ & 37.25 & 0.996874 & 0.006982 & 96 & 96/96 & 3 & 276 (97.87\%) \\
\bottomrule
\end{tabular}}
\end{table*}

\subsection{Transverse Watermark Scale}
\label{sec:supp-ablation-glyph-scale}

We vary the transverse Gaussian scale around the default 0.33 (Table~\ref{tab:supp-ablation-glyph-scale}). Target recognition stays between 95\% and 97\% for values from 0.20 to 0.40, then drops to 90\% at 0.45. Final non-target fidelity changes comparatively little. The 0.25 setting has 97\% overall accuracy but retains 95 of the fixed 96 baseline-recognized targets, illustrating why accuracy and paired retention are reported separately.

Changing the watermark scale also changes state $\mathbf{A}$. The stage references in Table~\ref{tab:supp-ablation-reference-states} therefore use separate $A/B$ values for each scale rather than a shared watermark-only baseline.

\begin{table*}[!tp]
\centering
\caption{Transverse watermark scale. Final fidelity uses all 9,600 non-target candidates. Target recognition accuracy uses 100 images and paired retention uses the fixed 96-image reference subset.}
\label{tab:supp-ablation-glyph-scale}
{\fontsize{8.8}{10.2}\selectfont
\setlength{\tabcolsep}{5.0pt}
\renewcommand{\arraystretch}{1.12}
\begin{tabular}{@{}rrrrrr@{}}
\toprule
Scale & PSNR $\uparrow$ & SSIM $\uparrow$ & LPIPS $\downarrow$ & Target acc. (\%) & Retained \\
\midrule
0.20 & 37.38 & 0.996877 & 0.007011 & 96 & 94/96 \\
0.25 & 37.34 & 0.996877 & 0.007009 & 97 & 95/96 \\
\rowcolor{gray!8} 0.33 & 37.25 & 0.996874 & 0.006982 & 96 & 96/96 \\
0.40 & 37.16 & 0.996869 & 0.006963 & 95 & 93/96 \\
0.45 & 37.09 & 0.996860 & 0.006958 & 90 & 89/96 \\
\bottomrule
\end{tabular}}
\end{table*}

\subsection{BO Loss Balance}
\label{sec:supp-ablation-bo-ratio}

We vary $r=\lambda_{\mathrm{non}}/\lambda_{\mathrm{tar}}$ while holding $\lambda_{\mathrm{non}}+\lambda_{\mathrm{tar}}=550$. The ratio 2.5 gives slightly better final non-target PSNR, SSIM, and LPIPS than the default ratio 10, together with 97\% rather than 96\% target recognition (Table~\ref{tab:supp-ablation-bo-ratio}). Increasing the ratio to 20 or 40 does not improve non-target fidelity and reduces recognition to 94\%. Target PSNR and SSIM decline as the ratio increases.

These results show that a larger non-target loss weight does not necessarily improve the achieved balance. Both ratios 2.5 and 5 retain 95/96 of the baseline-recognized targets.

\begin{table*}[!tp]
\centering
\caption{BO loss ratio with total weight fixed at 550. All values are from state $\mathbf{C}$. Non-target fidelity uses 9,600 candidates, target fidelity and accuracy use 100 images, and retention uses 96. The settings share states $\mathbf{A}$ and $\mathbf{B}$.}
\label{tab:supp-ablation-bo-ratio}
{\fontsize{8.4}{10.2}\selectfont
\setlength{\tabcolsep}{3.0pt}
\renewcommand{\arraystretch}{1.12}
\begin{tabular}{@{}rrrrrrrrr@{}}
\toprule
 & \multicolumn{3}{c}{All non-target candidates} & \multicolumn{3}{c}{Target view} & \multicolumn{2}{c}{Target recognition} \\
\cmidrule(lr){2-4}\cmidrule(lr){5-7}\cmidrule(lr){8-9}
$r$ & PSNR $\uparrow$ & SSIM $\uparrow$ & LPIPS $\downarrow$ & PSNR $\uparrow$ & SSIM $\uparrow$ & LPIPS $\downarrow$ & Acc. (\%) & Retained \\
\midrule
2.5 & 37.28 & 0.996935 & 0.006900 & 36.52 & 0.997746 & 0.007066 & 97 & 95/96 \\
5 & 37.27 & 0.996910 & 0.006930 & 36.49 & 0.997628 & 0.007169 & 97 & 95/96 \\
\rowcolor{gray!8} 10 & 37.25 & 0.996874 & 0.006982 & 36.42 & 0.997480 & 0.007311 & 96 & 96/96 \\
20 & 37.14 & 0.996819 & 0.007064 & 36.27 & 0.997308 & 0.007432 & 94 & 93/96 \\
40 & 36.99 & 0.996764 & 0.007107 & 36.10 & 0.997118 & 0.007551 & 94 & 92/96 \\
\bottomrule
\end{tabular}}
\end{table*}

\subsection{Rendering Resolution}
\label{sec:supp-ablation-resolution}

We render the same models and cameras at four resolutions while retaining the field of view and the default eight-view optimization. The models are not retrained. Every fidelity reference has the matching camera and resolution, and the recognition prompt adjusts the image and glyph pixel dimensions.

Target accuracy is 0\%, 96\%, 97\%, and 90\% at $128^2$, $256^2$, $512^2$, and $1024^2$, respectively (Table~\ref{tab:supp-ablation-resolution}). At $128^2$, the target PSNR is the highest despite the loss of letter recognition. Higher resolution also does not yield monotonic recognition gains. The fidelity changes describe rendering and sampling of the same models, so they should not be interpreted as evidence of improved optimization or used as a substitute for readability.

\begin{table*}[!tp]
\centering
\caption{Rendering resolution with fixed models, camera poses, and field of view. Final non-target fidelity aggregates 9,600 candidates, while target fidelity and accuracy use 100 images. Retention uses the fixed 96 baseline-recognized targets.}
\label{tab:supp-ablation-resolution}
{\fontsize{8.4}{10.2}\selectfont
\setlength{\tabcolsep}{3.0pt}
\renewcommand{\arraystretch}{1.12}
\begin{tabular}{@{}rrrrrrrrr@{}}
\toprule
 & \multicolumn{3}{c}{All non-target candidates} & \multicolumn{3}{c}{Target view} & \multicolumn{2}{c}{Target recognition} \\
\cmidrule(lr){2-4}\cmidrule(lr){5-7}\cmidrule(lr){8-9}
Resolution & PSNR $\uparrow$ & SSIM $\uparrow$ & LPIPS $\downarrow$ & PSNR $\uparrow$ & SSIM $\uparrow$ & LPIPS $\downarrow$ & Acc. (\%) & Retained \\
\midrule
$128^2$ & 37.88 & 0.997089 & 0.008311 & 39.3839 & 0.998080 & 0.008424 & 0 & 0/96 \\
\rowcolor{gray!8} $256^2$ & 37.25 & 0.996874 & 0.006982 & 36.4159 & 0.997480 & 0.007311 & 96 & 96/96 \\
$512^2$ & 37.73 & 0.996977 & 0.005385 & 37.9398 & 0.997621 & 0.004902 & 97 & 93/96 \\
$1024^2$ & 37.94 & 0.997358 & 0.004249 & 38.8597 & 0.997854 & 0.003667 & 90 & 87/96 \\
\bottomrule
\end{tabular}}
\end{table*}

\paragraph{Reference states for the fidelity comparisons.}
Table~\ref{tab:supp-ablation-reference-states} provides the $A/B$ measurements omitted from the final-state tables. Together, these tables specify the complete measured $\mathbf{A}/\mathbf{B}/\mathbf{C}$ progression without repeating the state-$\mathbf{C}$ values. Shared rows apply only to view selection, optimization-view count, and BO loss balance. Cavity length changes state $\mathbf{B}$, watermark scale changes state $\mathbf{A}$, and resolution changes the sampling and clean reference for both states.
The additional cavity-parameter and glyph-resolution
studies report their full-image $\mathbf{A}/\mathbf{B}/\mathbf{C}$ measurements
directly in Tables~\ref{tab:supp-ablation-bo-planar}
and~\ref{tab:supp-ablation-glyph-size}; their
configuration-specific states should not be inferred
from the shared reference rows.

\begin{table*}[!tp]
\centering
\caption{Reference-state fidelity, with each cell listing $A/B$. Shared target rows use 100 images and shared non-target rows use 9,600 candidates. These shared states apply to selection strategy, optimization-view count, and BO loss ratio. All length, scale, and resolution rows use 9,600 non-target candidates. Final $\mathbf{C}$ values appear in the corresponding tables above. Camera retreat has its own complete progression in Table~\ref{tab:supp-ablation-retreat}.}
\label{tab:supp-ablation-reference-states}
{\fontsize{8.5}{10.2}\selectfont
\setlength{\tabcolsep}{4.2pt}
\renewcommand{\arraystretch}{1.12}
\begin{tabular}{@{}llccc@{}}
\toprule
Sweep or population & Setting & PSNR $\uparrow$ ($A/B$) & SSIM $\uparrow$ ($A/B$) & LPIPS $\downarrow$ ($A/B$) \\
\midrule
Shared target & \multicolumn{1}{c}{All settings} & 36.00 / 35.95 & 0.997776 / 0.997539 & 0.006459 / 0.006877 \\
Shared non-target & \multicolumn{1}{c}{All settings} & 37.34 / 32.31 & 0.998323 / 0.995135 & 0.005245 / 0.008658 \\
\midrule
Length & $0.5\times$ & 37.34 / 34.15 & 0.998323 / 0.996599 & 0.005245 / 0.007229 \\
 & $0.625\times$ & 37.34 / 33.69 & 0.998323 / 0.996289 & 0.005245 / 0.007519 \\
 & $0.75\times$ & 37.34 / 33.24 & 0.998323 / 0.995953 & 0.005245 / 0.007853 \\
 & $0.875\times$ & 37.34 / 32.79 & 0.998323 / 0.995571 & 0.005245 / 0.008238 \\
\rowcolor{gray!8}  & $1\times$ & 37.34 / 32.31 & 0.998323 / 0.995135 & 0.005245 / 0.008658 \\
\midrule
Scale & 0.20 & 37.81 / 32.40 & 0.998383 / 0.995163 & 0.005278 / 0.008696 \\
 & 0.25 & 37.65 / 32.37 & 0.998363 / 0.995153 & 0.005267 / 0.008686 \\
\rowcolor{gray!8}  & 0.33 & 37.34 / 32.31 & 0.998323 / 0.995135 & 0.005245 / 0.008658 \\
 & 0.40 & 37.04 / 32.24 & 0.998281 / 0.995106 & 0.005231 / 0.008650 \\
 & 0.45 & 36.82 / 32.18 & 0.998248 / 0.995087 & 0.005228 / 0.008645 \\
\midrule
Resolution & $128^2$ & 34.95 / 31.10 & 0.997767 / 0.994545 & 0.009367 / 0.012660 \\
\rowcolor{gray!8}  & $256^2$ & 37.34 / 32.31 & 0.998323 / 0.995135 & 0.005245 / 0.008658 \\
 & $512^2$ & 39.94 / 33.32 & 0.999076 / 0.996030 & 0.002785 / 0.006283 \\
 & $1024^2$ & 41.52 / 33.71 & 0.999384 / 0.996772 & 0.001642 / 0.004964 \\
\bottomrule
\end{tabular}}
\end{table*}

\subsection{Focal-Length Sensitivity}
\label{sec:exp-focal-sensitivity}

We vary the focal-length multiplier
$\kappa\in\{1,1.5,2,2.5,3\}$ in
Eq.~\eqref{eq:ew-main-target-intrinsics}.
For a fixed image-space glyph, this changes its world-space footprint
and affects placement, parallax, and projection incidence.
The fidelity and projection sweeps below use distinct scene cohorts.

\paragraph{Rendering fidelity.}
Table~\ref{tab:exp-focal-fidelity} uses 50 scenes, with one target,
eight Seen cameras, and 32 dataset cameras per scene.
It compares the final state $\mathbf{C}$ with clean renderings.
The default $\kappa=2.5$ gives the highest target SSIM and the lowest
LPIPS in both target and dataset views among the tested settings.
Increasing $\kappa$ to $3$ raises dataset PSNR from 57.22 to 60.75~dB
but increases LPIPS in both groups.
Seen-view fidelity is highest at $\kappa=1$, showing that the preferred
setting depends on the evaluated camera distribution.
\paragraph{Projection incidence.}
A separate sweep uses 29 scenes and 928 dataset camera views.
Table~\ref{tab:exp-focal-projection} reports the incidence of nonempty
standalone projections from the watermark or sidewalls.
It falls from 426/928 (45.91\%) at $\kappa=1$ to 8/928 (0.86\%)
at $\kappa=2$ and 7/928 (0.75\%) at $\kappa=2.5$, then rises to
36/928 (3.88\%) at $\kappa=3$.
These results support $\kappa=2.5$ as a practical balance between
projection incidence and dataset-view perceptual fidelity.
The difference between $\kappa=2$ and $\kappa=2.5$ is only one view,
so this finite sweep does not establish a universal optimum.

\begin{table}[!htbp]
\centering
\caption{Focal-length sensitivity on 50 scenes, using a cohort distinct from the projection sweep. Fidelity compares the final $\mathbf{C}$ state with clean renderings. Each setting uses 50 Target, 400 Seen, and 1,600 Dataset images. LPIPS is reported in units of $10^{-2}$. The default $\kappa=2.5$ is shaded.}
\label{tab:exp-focal-fidelity}
\fontsize{8.5}{10}\selectfont
\setlength{\tabcolsep}{4pt}
\renewcommand{\arraystretch}{1.08}
\begin{tabular}{@{}rlrrr@{}}
\toprule
$\kappa$ & View & PSNR$\uparrow$ & SSIM$\uparrow$ & LPIPS$\downarrow$ \\
\midrule
1 & Target & 37.82 & 0.9974 & 0.925 \\
1.5 & Target & 37.58 & 0.9975 & 0.771 \\
2 & Target & 37.37 & 0.9975 & 0.782 \\
\rowcolor{gray!12}
2.5 & Target & 37.02 & 0.9976 & 0.754 \\
3 & Target & 36.54 & 0.9975 & 0.785 \\
\midrule
1 & Seen & 43.72 & 0.9969 & 0.887 \\
1.5 & Seen & 40.38 & 0.9940 & 1.378 \\
2 & Seen & 39.34 & 0.9925 & 1.653 \\
\rowcolor{gray!12}
2.5 & Seen & 39.38 & 0.9922 & 1.730 \\
3 & Seen & 39.06 & 0.9916 & 1.923 \\
\midrule
1 & Dataset & 35.21 & 0.9975 & 0.440 \\
1.5 & Dataset & 42.68 & 0.9996 & 0.083 \\
2 & Dataset & 51.71 & 0.9999 & 0.019 \\
\rowcolor{gray!12}
2.5 & Dataset & 57.22 & 1.000 & 0.007 \\
3 & Dataset & 60.75 & 1.000 & 0.009 \\
\bottomrule
\end{tabular}
\end{table}
\setcounter{topnumber}{3}
\setcounter{bottomnumber}{2}
\setcounter{totalnumber}{5}
\setcounter{dbltopnumber}{3}
\renewcommand{\topfraction}{0.95}
\renewcommand{\bottomfraction}{0.85}
\renewcommand{\textfraction}{0.05}
\renewcommand{\dbltopfraction}{0.95}
\setlength{\textfloatsep}{9pt plus 2pt minus 2pt}
\setlength{\dbltextfloatsep}{9pt plus 2pt minus 2pt}
\setlength{\floatsep}{7pt plus 2pt minus 2pt}
\setlength{\dblfloatsep}{7pt plus 2pt minus 2pt}
\setlength{\intextsep}{7pt plus 2pt minus 2pt}

\subsection{Cavity Optimization and Glyph Resolution}
\label{sec:supp-bo-size-discussion}
\paragraph{Cavity parameters and planar control.}
Table~\ref{tab:supp-ablation-bo-planar} compares frozen
sidewalls, appearance-only optimization, geometry/opacity
optimization, and full BO, using the same calibrated
initialization, watermark, target camera, and training views.
Full BO achieves the highest all-candidate PSNR among these
four configurations, improving from 32.31 to 37.25 dB.
Geometry/opacity optimization yields slightly higher SSIM
than full BO (0.996917 versus 0.996874), so joint optimization
does not dominate every metric.
The planar control replaces the watermark and sidewalls
with 540 learnable coplanar Gaussians at the watermark depth,
without longitudinal sidewalls. For this control, A retains the
original learned watermark, whereas $\mathbf{B}$ and $\mathbf{C}$ use the planar
replacement before and after optimization, respectively.
Although it achieves higher all-candidate PSNR (40.83 dB),
target recognition decreases from 96/100 at A to 76/100 at C,
with 75/96 originally readable targets retained.
At $0.25^\circ$ and $0.5^\circ$, its stage-C recognition rates
are 44/600 and 12/600, respectively, and its conditional
suppression rates are 79.43\% (166/209) and 93.81\% (182/194).
Thus, higher image fidelity alone does not establish better
view-selective watermarking.

\begin{table*}[!tp]
\centering
\setlength{\abovecaptionskip}{3pt}
\setlength{\belowcaptionskip}{4pt}
\caption{Cavity-parameter and planar-patch ablation on the same
100 scenes. The shaded row denotes the full-BO baseline.
The planar control replaces the watermark and sidewalls with
540 coplanar learnable Gaussians at the watermark depth.
PSNR is reported in dB.}
\label{tab:supp-ablation-bo-planar}
{\fontsize{8.7}{10.5}\selectfont
\setlength{\tabcolsep}{5pt}
\renewcommand{\arraystretch}{1.08}

\textbf{(a) Stage-C local metrics: $8\times8$ crops}
\par\vspace{2pt}
\begin{tabular}{@{}lrrrrrr@{}}
\toprule
Configuration
& \multicolumn{2}{c}{Target}
& \multicolumn{2}{c}{Seen}
& \multicolumn{2}{c}{Unseen} \\
& PSNR & SSIM & PSNR & SSIM & PSNR & SSIM \\
\midrule
Frozen sidewalls
& 5.95 & 0.079008 & 7.52 & 0.148855 & 6.89 & 0.113921 \\
Appearance only
& 5.95 & 0.079008 & 16.43 & 0.556990 & 6.92 & 0.097874 \\
Geometry + opacity
& 6.47 & 0.068696 & 13.32 & 0.228170 & 7.25 & 0.077045 \\
\rowcolor{gray!8}
Full BO (baseline)
& 6.61 & 0.067248 & 19.40 & 0.597582 & 8.01 & 0.084860 \\
Planar patch (540 GS)
& 8.79 & 0.097215 & 19.37 & 0.489652 & 9.40 & 0.115721 \\
\bottomrule
\end{tabular}

\par\vspace{4pt}
\textbf{(b) Stage-C full-image metrics}
\par\vspace{2pt}
\begin{tabular}{@{}lrrrrrr@{}}
\toprule
Configuration
& \multicolumn{2}{c}{Target}
& \multicolumn{2}{c}{Seen}
& \multicolumn{2}{c}{Unseen} \\
& PSNR & SSIM & PSNR & SSIM & PSNR & SSIM \\
\midrule
Frozen sidewalls
& 35.95 & 0.997539 & 23.57 & 0.972371 & 36.18 & 0.997205 \\
Appearance only
& 35.55 & 0.997006 & 35.06 & 0.986582 & 35.86 & 0.996685 \\
Geometry + opacity
& 36.17 & 0.997455 & 33.70 & 0.987729 & 37.28 & 0.997752 \\
\rowcolor{gray!8}
Full BO (baseline)
& 36.42 & 0.997480 & 38.60 & 0.991337 & 37.14 & 0.997377 \\
Planar patch (540 GS)
& 38.89 & 0.998142 & 49.71 & 0.999304 & 40.50 & 0.998653 \\
\bottomrule
\end{tabular}

\par\vspace{4pt}
\textbf{(c) All 9,600 non-target candidate views: full-image $\mathbf{A}/\mathbf{B}/\mathbf{C}$}
\par\vspace{2pt}
\begin{tabular}{@{}lrrrrrr@{}}
\toprule
Configuration
& \multicolumn{2}{c}{A}
& \multicolumn{2}{c}{B}
& \multicolumn{2}{c}{C} \\
& PSNR & SSIM & PSNR & SSIM & PSNR & SSIM \\
\midrule
Frozen sidewalls
& 37.34 & 0.998323 & 32.31 & 0.995135 & 32.31 & 0.995135 \\
Appearance only
& 37.34 & 0.998323 & 32.31 & 0.995135 & 35.79 & 0.995843 \\
Geometry + opacity
& 37.34 & 0.998323 & 32.31 & 0.995135 & 36.84 & 0.996917 \\
\rowcolor{gray!8}
Full BO (baseline)
& 37.34 & 0.998323 & 32.31 & 0.995135 & 37.25 & 0.996874 \\
Planar patch (540 GS)
& 37.34 & 0.998323 & 37.27 & 0.998310 & 40.83 & 0.998707 \\
\bottomrule
\end{tabular}

\par\vspace{4pt}
\textbf{(d) Recognition: accuracy and conditional suppression (\%)}
\par\vspace{2pt}
\begin{tabular}{@{}llrrr@{}}
\toprule
Configuration & Rotation & Acc. $\mathbf{A}$ & Acc. $\mathbf{C}$ & Suppression \\
\midrule
Planar patch (540 GS) & $0^\circ$
& 96.00 (96/100) & 76.00 (76/100)
& 21.88 (21/96)$^{\dagger}$ \\
Planar patch (540 GS) & $0.25^\circ$
& 34.83 (209/600) & 7.33 (44/600)
& 79.43 (166/209) \\
Planar patch (540 GS) & $0.5^\circ$
& 32.33 (194/600) & 2.00 (12/600)
& 93.81 (182/194) \\
\bottomrule
\end{tabular}

\par\vspace{2pt}
\begin{minipage}{0.98\textwidth}
\raggedright
$\dagger$ At $0^\circ$, suppression denotes target readability
loss, not hiding success.
Conditional suppression counts A-correct images that are
no longer correctly recognized at C, including wrong-letter
responses. Panel (a) uses 100 Target, 752 Seen, and 6,878
Unseen valid crops for every configuration. Crops require a
valid projected watermark center and a complete $8\times8$
region, with no padding or recentering; the same eligible
cameras are used across $\mathbf{A}/\mathbf{B}/\mathbf{C}$. Panel (b) uses all 100 Target,
800 Seen, and 8,800 Unseen views.
\end{minipage}
}
\end{table*}

\paragraph{Glyph resolution.}
Table~\ref{tab:supp-ablation-glyph-size} compares
$8\times8$, $16\times16$, and $24\times24$ versions of IBM
glyph A. Larger glyphs use integer nearest-neighbor
replication and proportionally denser Gaussian constructions.
They repeat size-aware placement, adaptive color optimization,
and view selection while retaining the $32\times32$
corner-search window; this is an end-to-end configuration
ablation rather than a fixed-camera comparison.
We report full-image fidelity for this size comparison:
100 Target, 800 Seen, and 8,800 Unseen views per configuration,
with all 9,600 non-target candidates included in the pooled metrics.
Each configuration uses its own $\mathbf{A}/\mathbf{B}/\mathbf{C}$ images at matched cameras;
the larger-glyph A/B values are not shared with the baseline.
Stage-C all-candidate PSNR decreases from 37.25 dB for the
$8\times8$ baseline to 32.20 and 29.16 dB for the larger
variants. Nevertheless, target recognition remains 99/100
at both $\mathbf{A}$ and $\mathbf{C}$ for $16\times16$, and 100/100 at both stages
for $24\times24$; every originally readable target is retained.
At $0.25^\circ$, conditional suppression reaches
100\% (234/234) and 99.58\% (235/236), respectively.
At $0.5^\circ$, both variants have zero correct $\mathbf{C}$-stage
recognitions among 600 images, suppressing all 215 and 214
A-readable images, respectively.
These results show that larger glyphs remain readable at
the target and strongly suppress correct recognition at
the tested rotations, at the cost of greater image deviation.

\begin{table*}[!tp]
\centering
\setlength{\abovecaptionskip}{3pt}
\setlength{\belowcaptionskip}{4pt}
\caption{Glyph-resolution ablation on the same 100 scenes.
Larger glyphs use size-aware placement and proportionally
denser Gaussian constructions; this is not a fixed-camera
comparison. The shaded row denotes the original $8\times8$
baseline. PSNR is reported in dB.}
\label{tab:supp-ablation-glyph-size}
{\fontsize{8.7}{10.5}\selectfont
\setlength{\tabcolsep}{5pt}
\renewcommand{\arraystretch}{1.08}

\textbf{(a) Stage-C full-image metrics}
\par\vspace{2pt}
\begin{tabular}{@{}lrrrrrr@{}}
\toprule
Configuration
& \multicolumn{2}{c}{Target}
& \multicolumn{2}{c}{Seen}
& \multicolumn{2}{c}{Unseen} \\
& PSNR & SSIM & PSNR & SSIM & PSNR & SSIM \\
\midrule
\rowcolor{gray!8}
$8\times8$ (baseline)
& 36.42 & 0.997480 & 38.60 & 0.991337 & 37.14 & 0.997377 \\
$16\times16$
& 31.11 & 0.994274 & 37.45 & 0.989706 & 31.94 & 0.994004 \\
$24\times24$
& 28.06 & 0.991845 & 35.07 & 0.989017 & 28.88 & 0.991503 \\
\bottomrule
\end{tabular}

\par\vspace{4pt}
\textbf{(b) All 9,600 non-target candidate views: full-image $\mathbf{A}/\mathbf{B}/\mathbf{C}$}
\par\vspace{2pt}
\begin{tabular}{@{}lrrrrrr@{}}
\toprule
Configuration
& \multicolumn{2}{c}{$\mathbf{A}$}
& \multicolumn{2}{c}{$\mathbf{B}$}
& \multicolumn{2}{c}{$\mathbf{C}$} \\
& PSNR & SSIM & PSNR & SSIM & PSNR & SSIM \\
\midrule
\rowcolor{gray!8}
$8\times8$ (baseline)
& 37.34 & 0.998323 & 32.31 & 0.995135 & 37.25 & 0.996874 \\
$16\times16$
& 32.24 & 0.996210 & 29.65 & 0.992155 & 32.20 & 0.993646 \\
$24\times24$
& 29.15 & 0.994022 & 27.46 & 0.989461 & 29.16 & 0.991296 \\
\bottomrule
\end{tabular}

\par\vspace{4pt}
\textbf{(c) Recognition: accuracy and conditional suppression (\%)}
\par\vspace{2pt}
\begin{tabular}{@{}llrrr@{}}
\toprule
Configuration & Rotation & Acc. $\mathbf{A}$ & Acc. $\mathbf{C}$ & Suppression \\
\midrule
\rowcolor{gray!8}
$8\times8$ (baseline) & $0^\circ$
& --- & 96.00 (96/100) & --- \\
$16\times16$ & $0^\circ$
& 99.00 (99/100) & 99.00 (99/100)
& 0.00 (0/99)$^{\dagger}$ \\
$16\times16$ & $0.25^\circ$
& 39.00 (234/600) & 0.00 (0/600)
& 100.00 (234/234) \\
$16\times16$ & $0.5^\circ$
& 35.83 (215/600) & 0.00 (0/600)
& 100.00 (215/215) \\
$24\times24$ & $0^\circ$
& 100.00 (100/100) & 100.00 (100/100)
& 0.00 (0/100)$^{\dagger}$ \\
$24\times24$ & $0.25^\circ$
& 39.33 (236/600) & 0.17 (1/600)
& 99.58 (235/236) \\
$24\times24$ & $0.5^\circ$
& 35.67 (214/600) & 0.00 (0/600)
& 100.00 (214/214) \\
\bottomrule
\end{tabular}

\par\vspace{2pt}
\begin{minipage}{0.98\textwidth}
\raggedright
$\dagger$ At $0^\circ$, suppression denotes target readability
loss, not hiding success.
The baseline target-$\mathbf{C}$ accuracy is reused; dashes denote
matched recognition quantities not reported in this scan.
Conditional suppression is evaluated only on A-correct
images with matched $\mathbf{C}$ responses, including wrong-letter
responses as incorrect. Panel (a) uses 100 Target, 800 Seen,
and 8,800 Unseen full images per configuration; panel (b)
pools all 9,600 non-target candidates. No local-crop
validity filter is applied to these full-image metrics.
\end{minipage}
}
\end{table*}

\paragraph{Recognition protocol and interpretation.}
Recognition uses the requested model \texttt{gpt-6-luna},
high reasoning effort and image detail, the same prompt, and an IBM font reference, without
the clean image or ground-truth letter.
Each rotation magnitude pools yaw, pitch, and diagonal
rotations in both directions over 100 scenes, giving 600
matched positions. These positions are not restricted to
Unseen cameras.
Conditional suppression is the fraction of A-correct images
that are no longer correctly recognized at C, including
wrong-letter responses; at the target camera, the same
quantity represents readability loss.

\begin{table}[!htbp]
\centering
\setlength{\abovecaptionskip}{3pt}
\setlength{\belowcaptionskip}{4pt}
\caption{Projection sensitivity in a separate 29-scene focal-length sweep, with 32 original cameras per scene and 928 cameras in total. Counts indicate nonempty standalone projections. Sidewall counts equal the Either counts at every setting, and parentheses give Either rates in percent. The Scenes column counts affected scenes out of 29. This cohort is distinct from the 50-scene fidelity experiment in Table~\ref{tab:exp-focal-fidelity}. The default $\kappa=2.5$ is shaded and has the lowest observed incidence.}
\label{tab:exp-focal-projection}
\fontsize{8.5}{10}\selectfont
\setlength{\tabcolsep}{4pt}
\renewcommand{\arraystretch}{1.08}
\begin{tabular}{@{}rrrr@{}}
\toprule
$\kappa$ & Watermark & Either (\%) & Scenes \\
\midrule
1 & 393/928 & 426/928 (45.91) & 28/29 \\
1.5 & 78/928 & 79/928 (8.51) & 6/29 \\
2 & 7/928 & 8/928 (0.86) & 1/29 \\
\rowcolor{gray!12}
2.5 & 6/928 & 7/928 (0.75) & 1/29 \\
3 & 36/928 & 36/928 (3.88) & 2/29 \\
\bottomrule
\end{tabular}
\end{table}
\subsection{Additional Visibility Diagnostics}
\label{sec:visibility_readability}

\paragraph{Angular projection analysis.}
We report an earlier-configuration diagnostic on 117 optimized
scenes. For 117 optimized scenes, we orbit cameras around the Gaussian scene
center and measure the fraction of views with no projection from the
specified Gaussian group.
Table~\ref{tab:projection_absence} reports watermark projection absence
of 86.0\% at $2^{\circ}$ and 98.9\% at $5^{\circ}$.
Joint absence of the watermark and cavity reaches 90.6\% at
$2.7^{\circ}$ and 98.3\% at $5^{\circ}$.
These measurements describe the angular extent of the inserted structure.
They do not determine whether a glyph still in the frame is readable
or whether its disappearance is caused by cavity occlusion.

\begin{table}[!htbp]
  \centering
\setlength{\abovecaptionskip}{3pt}
\setlength{\belowcaptionskip}{4pt}
  \caption{Projection absence across 117 optimized scenes.
  Each entry is the percentage of evaluated views at the
  specified angular offset for which the indicated Gaussian
  group has no projection. This statistic is distinct from
  letter-recognition accuracy.}
  \label{tab:projection_absence}
  \small
  \setlength{\tabcolsep}{5pt}
  \begin{tabular}{@{}lcc@{}}
    \toprule
    Gaussian group & Offset & No projection (\%) \\
    \midrule
    Watermark & $2.0^{\circ}$ & 86.0 \\
    Watermark & $5.0^{\circ}$ & 98.9 \\
    Watermark + cavity & $2.7^{\circ}$ & 90.6 \\
    Watermark + cavity & $5.0^{\circ}$ & 98.3 \\
    \bottomrule
  \end{tabular}
\end{table}

\paragraph{Matched-view readability.}
A separate readability evaluation uses ten selected scenes and Kimi3
as a vision-language model evaluator.
Its 1,080 images comprise 540 matched camera settings, each rendered
with the watermark alone and with the optimized cavity.
The cameras cover nine angular offsets between $0.5^{\circ}$ and
$6^{\circ}$ under three rotation modes.
This cohort and its recognition outcomes are distinct from the
117-scene projection statistics above.

Without the cavity, 436 of the 540 images are pixel-identical to their
clean counterparts and 104 contain a visible glyph.
Within these 104 views, 71 glyphs are judged clearly recognizable,
29 tentatively recognizable, and four unrecognizable.
Counting clear and tentative judgments as recognizable gives
$100/104=96.15\%$.
With the optimized cavity at the same camera settings, 103 of these
104 glyphs are judged unrecognizable, leaving one recognizable case
($1/104=0.96\%$), as shown in Table~\ref{tab:vlm_readability}.
This paired result supports reduced off-target readability where the
watermark-only rendering exposes a glyph.
The criterion does not require correct transcription of the intended
letter, and unreadability does not establish complete visual concealment.
A residual glyph or an occluding wall can still produce an artifact.

\begin{table}[!htbp]
  \centering
\setlength{\abovecaptionskip}{3pt}
\setlength{\belowcaptionskip}{4pt}
  \caption{VLM-based readability assessment on matched camera
  views. Entries are counts with percentages in parentheses.
  The first panel uses all 540 camera settings per condition.
  The second uses only the 104 settings with a visible glyph
  in the watermark-only rendering.
  Recognizable includes both clear and tentative judgments.
  It does not imply correct transcription of the intended
  letter.}
  \label{tab:vlm_readability}
  \small
  \setlength{\tabcolsep}{4pt}
  \begin{tabular}{@{}lcc@{}}
    \toprule
    Measure & No cavity & With cavity \\
    \midrule
    \multicolumn{3}{@{}l}{All sampled views ($N=540$)} \\
    Pixel-identical to clean
      & 436 (80.74\%) & 429 (79.44\%) \\
    \midrule
    \multicolumn{3}{@{}l}{
      Paired glyph-containing views ($N=104$)} \\
    Recognizable glyph
      & 100 (96.15\%) & 1 (0.96\%) \\
    Unrecognizable glyph
      & 4 (3.85\%) & 103 (99.04\%) \\
    \bottomrule
  \end{tabular}
\end{table}
\paragraph{Camera-translation protocol.}
As a separate experiment, we evaluate camera-translation sensitivity on 100 randomly selected MVSplat/RE10K scenes using IBM glyph A, adaptive watermark colors, per-scene maximum focal lengths scaled by $2.5$, and cavities optimized for 600 steps.
Stage A contains the original scene and the learned watermark
without a cavity; stage C adds the optimized cavity.
This experiment uses a different cohort and recognition
protocol from the preceding readability evaluation.

Starting from the target camera $(R_\star,t_\star)$, we
translate along each signed local camera axis while keeping
orientation and intrinsics fixed:
\begin{equation}
t_{a,\epsilon}(s)
=
t_\star+
\epsilon\frac{s z_m}{f_x^\star}R_\star e_a,
\qquad
a\in\{x,y,z\},\quad \epsilon\in\{-1,+1\},
\end{equation}
where $z_m=0.33$ is the cavity midpoint depth and
$f_x^\star$ is the target horizontal focal length in pixels
at $256\times256$ resolution.
The parameter $s$ denotes a reference-depth pixel-equivalent
step, rather than the actual displacement of every projected
primitive. The implementation uses $f_x^\star$ for all
three axes. At depth $z_m$, the horizontal displacement
magnitude is $s$, while the vertical displacement magnitude
is $s f_y^\star/f_x^\star$.
Axial translation changes depth and does not produce a
uniform image shift.

\paragraph{Translation-induced projection disappearance.}
We scan integer steps $s=0,\ldots,64$ in all six directions.
The frozen watermark and optimized cavity are rasterized
independently to measure their standalone support, without
occlusion by the original scene.
Table~\ref{tab:supp-translation-projection} reports the
first sampled zero-support steps and the fraction of scenes
with an observed disappearance by $s=64$.

Across transverse directions, first observed disappearance
occurs at $s=21$--$35$ for the watermark and
$s=26$--$43$ for the joint structure, conditional on
disappearance being observed.
Neither the watermark nor the cavity disappears under
either axial direction within this scan.
These measurements describe geometric projection support;
they do not establish unreadability or successful cavity
occlusion.

\begin{table}[!htbp]
\centering
\setlength{\abovecaptionskip}{3pt}
\setlength{\belowcaptionskip}{4pt}
\caption{Translation-induced projection disappearance on
100 scenes per direction. Ranges contain the first sampled
zero-support steps among scenes with an observed
disappearance; rates use all 100 scenes.
Joint absence requires both the watermark and optimized
cavity to have empty standalone support.
Dashes indicate no observed disappearance within
$s\in[0,64]$.}
\label{tab:supp-translation-projection}
{\fontsize{8.7}{10.5}\selectfont
\setlength{\tabcolsep}{4pt}
\renewcommand{\arraystretch}{1.08}
\begin{tabular}{@{}lrrrr@{}}
\toprule
& \multicolumn{2}{c}{Watermark}
& \multicolumn{2}{c}{Joint} \\
Direction & Range & Rate (\%) & Range & Rate (\%) \\
\midrule
$-x$ & 24--31 & 31 & 27--38 & 31 \\
$+x$ & 21--29 & 69 & 27--38 & 69 \\
$-y$ & 24--35 & 51 & 27--43 & 51 \\
$+y$ & 22--33 & 49 & 26--42 & 49 \\
$-z$ & --- & 0 & --- & 0 \\
$+z$ & --- & 0 & --- & 0 \\
\bottomrule
\end{tabular}
}
\par\vspace{2pt}
\begin{minipage}{\linewidth}
\footnotesize
A first-zero sample $s_0$ brackets the preceding transition
in $(s_0-1,s_0]$.
Disappearance is not assumed to be monotonic:
one $+x$ trajectory exhibits subsequent cavity-support
reappearance.
\end{minipage}
\end{table}

\paragraph{Matched recognition under translation.}
We evaluate recognition at
$s\in\{10,16,24,32,48,64\}$, giving 600 matched camera
settings per step and 3,600 A/C pairs in total.
Recognition uses \texttt{gpt-6-luna}, high reasoning effort
and image detail, the same prompt,
and an IBM font reference, without the clean image or
ground-truth letter.

Conditional suppression is the fraction of A-correct
images that are no longer correctly recognized at C,
including wrong-letter responses.
Strict hiding counts only ``none'' or ``uncertain''
responses within the same A-correct subset.
These definitions differ from the clear-or-tentative
readability criterion used in
Table~\ref{tab:vlm_readability}.
Table~\ref{tab:supp-translation-recognition} reports
recognition outcomes over all sampled positions.

\begin{table}[!tp]
\centering
\setlength{\abovecaptionskip}{3pt}
\setlength{\belowcaptionskip}{4pt}
\caption{Recognition under camera translation on 100
MVSplat/RE10K scenes.
Accuracy uses all 600 sampled positions at each step,
covering six translation directions per scene.
Suppression and strict hiding are conditional on correct
recognition at A; counts are given in parentheses.
Suppression includes wrong-letter responses, whereas
strict hiding includes only ``none'' or ``uncertain''.
The same cameras are used across $\mathbf{A}$/$\mathbf{C}$.}
\label{tab:supp-translation-recognition}
{\fontsize{8.7}{10.5}\selectfont
\setlength{\tabcolsep}{1pt}
\renewcommand{\arraystretch}{1.08}
\begin{tabular}{@{}rrrrrr@{}}
\toprule
$s$ & $N$
& Acc. $\mathbf{A}$ (\%) & Acc. $\mathbf{C}$ (\%)
& Suppression (\%) & Strict hiding (\%) \\
\midrule
10 & 600 & 89.83 & 13.33
& 85.34 (460/539) & 79.22 (427/539) \\
16 & 600 & 74.33 & 3.33
& 95.52 (426/446) & 90.81 (405/446) \\
24 & 600 & 55.67 & 1.17
& 97.90 (327/334) & 95.51 (319/334) \\
32 & 600 & 59.17 & 0.17
& 100.00 (355/355) & 98.87 (351/355) \\
48 & 600 & 56.33 & 0.00
& 100.00 (338/338) & 99.11 (335/338) \\
64 & 600 & 47.83 & 0.00
& 100.00 (287/287) & 98.26 (282/287) \\
\bottomrule
\end{tabular}
}
\end{table}

Across all sampled positions, $\mathbf{C}$-stage accuracy decreases
from 13.33\% at $s=10$ to 3.33\% at $s=16$,
1.17\% at $s=24$, and 0.17\% at $s=32$,
reaching zero at $s=48$ and $s=64$.
Conditional suppression reaches 100\% at
$s=32,48,64$, while strict hiding remains lower
because wrong-letter responses are excluded.
At $s=32$, the single correctly recognized $\mathbf{C}$-stage image
was not correctly recognized at $\mathbf{A}$ and therefore lies
outside the A-correct subset used to compute conditional
suppression.
These observations concern the tested translations and
camera protocol; they do not establish concealment at
arbitrary translations, throughout a continuous displacement
interval, or disappearance of the inserted structure.


\section{LLM Usage Statement}
We used GPT-6 Astra to help revise the code for our LLM-based watermark recognition evaluation pipeline and to edit the manuscript for clarity and readability.
For code development, it assisted in implementing parts of the experimental scripts, as well as scripts for generating tables and figures. We used GPT-6 Luna to assess watermark recognition and Kimi3 to assess glyph readability in the supplementary matched-view experiment. The authors reviewed the code and manuscript revisions and take full responsibility for the evaluation procedure and all reported results.

\end{document}